\documentclass{article}

\usepackage{microtype}
\usepackage{graphicx}
\usepackage{subcaption}
\usepackage{booktabs} 
\usepackage{natbib}
\usepackage{xurl}

\usepackage{hyperref}

\usepackage[accepted]{icml2026}

\usepackage{amsmath}
\usepackage{amssymb}
\usepackage{mathtools}
\usepackage{amsthm}

\usepackage[capitalize,noabbrev, nameinlink]{cleveref}
\makeatletter
\AddToHook{cmd/appendix/before}{\def\cref@section@alias{appendix}\def\cref@subsection@alias{appendix}}
\makeatother

\usepackage{enumitem}
\usepackage{array}

\theoremstyle{plain}

\theoremstyle{definition}

\theoremstyle{remark}

\usepackage[disable,textsize=tiny]{todonotes}

\icmltitlerunning{Hippocampal Explicit Memory Is the Cornerstone for AGI}

\begin{document}

\twocolumn[
  \icmltitle{Position: Hippocampal Explicit Memory Is the Cornerstone for AGI}



  \icmlsetsymbol{equal}{*}

  \begin{icmlauthorlist}
    \icmlauthor{Sangjun Park}{ut,cail}
  \end{icmlauthorlist}

  \icmlaffiliation{ut}{Department of Computer Science, University of Texas at Austin, TX, USA}
  \icmlaffiliation{cail}{Cognizant AI Labs, San Francisco, CA, USA}

  \icmlcorrespondingauthor{Sangjun Park}{sangjun@cs.utexas.edu}

  \icmlkeywords{Machine Learning, ICML, Hippocampal Memory, Explicit Memory, Human-Level AI, AGI, LLMs, NeuroAI}

  \vskip 0.3in
]



\printAffiliationsAndNotice{}  

\begin{abstract}
    Large Language Models (LLMs) have demonstrated remarkable capabilities across various tasks, raising expectations for Artificial General Intelligence (AGI). This position paper argues that integrating explicit memory is the cornerstone for advancing LLMs toward AGI. The key reason is that the underlying learning mechanism of LLMs is highly analogous to human implicit memory. However, higher-order cognitive functions necessary for AGI, such as long-term strategic planning, metacognition, and symbolic reasoning, heavily rely on hippocampal explicit memory and cannot arise solely from implicit statistical learning. Drawing on findings from neuroscience, I advance this perspective and complement it with computational requirements for artificial explicit memory systems, hoping to foster further research and lay the groundwork for explicit memory integration.
\end{abstract}

\section{Introduction}

Large Language Models (LLMs) have recently achieved remarkable success in the field of natural language processing \citep{comanici2025gemini25pushingfrontier, singh2026openaigpt5card}, opening new possibilities in artificial intelligence \citep{qwen25, deepseek-r1}. Trained on vast amounts of text data, these models comprehend context and demonstrate near-human performance in tasks such as writing, question-answering, code generation, and conversational assistance \citep{emergent, bigbench, helm}. Notably, advanced models like GPT-5, utilizing hundreds of billions of parameters, can solve complex problems and have practical applications across various domains \citep{llm-healthcare, llm-poem, llm-law}. Some researchers even argue that LLMs can be considered the initial stages of Artificial General Intelligence (AGI), due to their ability to understand and address intricate issues \citep{ms-gpt4-agi}.

However, despite their impressive performance, LLMs still face significant challenges such as hallucination, difficulties in planning, and limitations in logical reasoning \citep{llm-not-abstract-reason, tree-of-thought, llm-creativity, llm-cog-failures, llm-still-cant-plan}.
Since LLMs cannot store and use information dynamically over long periods, memory has always been considered a notable weakness \citep{generative-ai-vs-agi, how-far-agi-llm, memorybank, levels-of-agi}.
I regard memory as the key point for addressing these issues and also for developing higher-order capabilities essential for AGI \citep{lake}, such as dynamic learning, reflection, and metacognition.
While there are various discussions regarding the definition of AGI, I consider Human-Level AI to be the defining criterion for AGI: an AI system with general, human-level ability to learn, reason, and apply knowledge across all cognitive tasks and domains.
Although memory is commonly perceived as merely information storage, in reality, it is an integral component of all learning processes in humans \citep{human-memory-atkinson, acquisition-cog-skill}. Constraints on memory are directly linked to limitations in cognitive ability, highlighting its pivotal role in the evolution of artificial intelligence.

In this paper, I argue that the fundamental learning mechanism of LLMs can be compared to learning based on implicit memory in biological systems. I also suggest that the cognitive functions of LLMs are restricted to implicit learning. Based on this perspective, \textbf{I claim that realizing AGI should involve the integration of explicit memory.}

\begin{figure*}[t!]
    \centerline{\includegraphics[width=\textwidth]{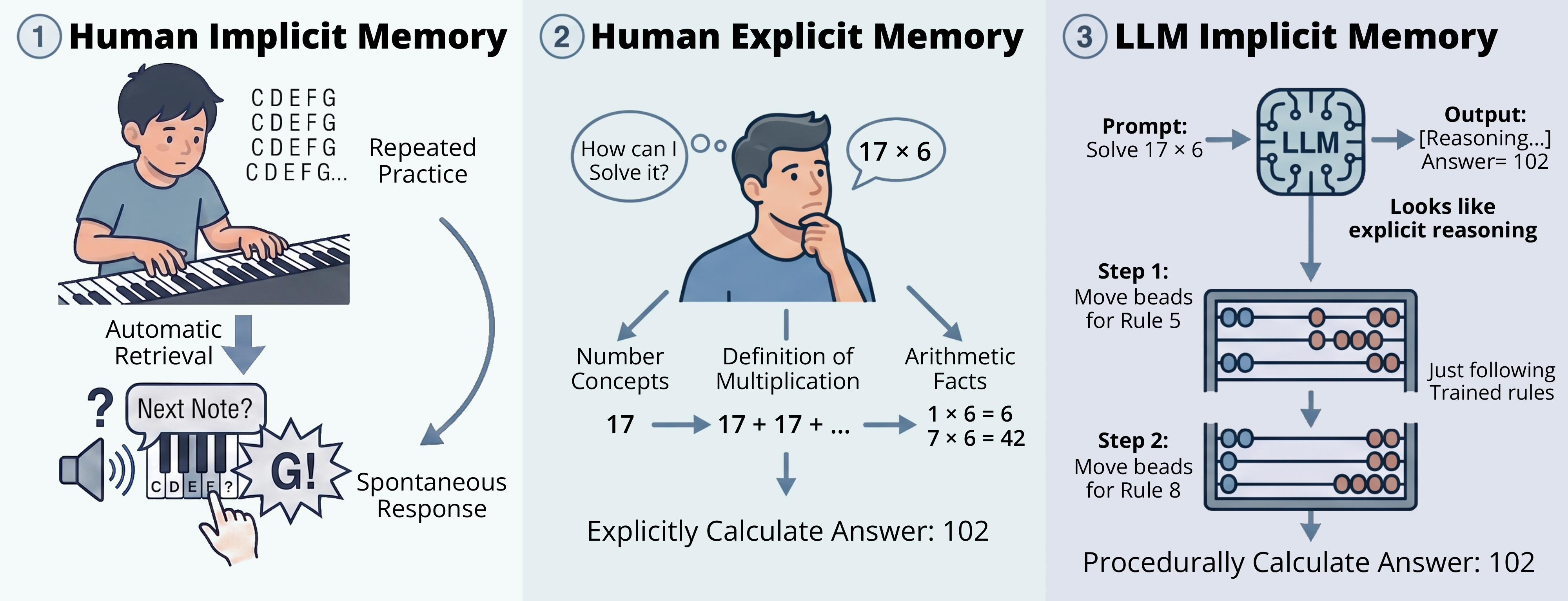}}
    \caption{Comparison of three memory and learning paradigms. \textbf{(1)} Human implicit
    memory forms through repeated practice and produces automatic, spontaneous
    responses (e.g., predicting the next note in a rehearsed melody).
    \textbf{(2)} Human explicit memory supports conscious, knowledge-based
    problem solving: $17 \times 6$ is decomposed via number concepts, the
    definition of multiplication, and arithmetic facts to explicitly compute
    $102$. \textbf{(3)} An LLM, given the same prompt, emits a reasoning-like
    trace ending in $102$. Although the output looks like explicit
    reasoning, the underlying process is closer to operating an abacus: the
    correct answer can be produced automatically by just following learned manipulation rules (rule indices are illustrative), with no semantic understanding of the numbers or the procedure.}
    \label{fig:memory-system-comparison}
\end{figure*}

To support this claim, the remainder of this paper is organized as follows. \cref{sec:explicit-and-implicit-mems} reviews the fundamental concepts and mechanisms of human explicit and implicit memory systems. \cref{sec:nature-of-learning-llms} analyzes the underlying learning processes of current LLMs, illustrating their structural and functional resemblance to implicit memory. \cref{sec:functional-specificity-mem} explores the distinct cognitive functions supported by each memory system, to clarify the operational boundaries of each memory. Building upon these insights, \cref{sec:need-for-explicit-memory} details the specific limitations of current LLMs and highlights why explicit memory is essential for higher-order cognition. \cref{sec:computational-requirements} then formally defines the computational requirements for developing an artificial explicit memory system.
\cref{sec:alternative-views} examines alternative perspectives, \cref{sec:discussion} discusses broader implications, and \cref{sec:conclusion} concludes the paper with directions for future research.
In the Appendices, \cref{sec:recent-progress} reviews recent advancements in neural memory. \cref{sec:further-discussions} further discusses theoretical and practical considerations such as substrate independence and causality. Finally, \cref{sec:evidence-absence-exmem} presents empirical examples demonstrating the absence of explicit memory in current LLMs.

\section{Explicit and Implicit Memories}
\label{sec:explicit-and-implicit-mems}

LLMs naturally possess memory, and understanding the memorization process in LLMs can be greatly enriched by drawing parallels with biological systems.
Human memory is structured to store and utilize multiple types of information \citep{Tulving-episodic-and-semantic, working-mem}, enabling various cognitive functions.
This section provides the basic concepts and operation mechanisms in each memory system.
Please refer to \cref{fig:memory-system-comparison} for illustrative examples of explicit and implicit memory systems in humans and LLMs.

\subsection{Basic Concepts}

\textbf{Explicit Memory} Explicit memory, also known as declarative memory, handles information related to facts, events, or experiences \citep{tulving-memory-consc}.
Explicit memory is typically divided into two main categories: episodic memory and semantic memory.
Episodic memory preserves personal experiences tied to specific times and places \citep{tulving-epi-mem}, so it is closely related to autobiographical memory and is instrumental in reconstructing past experiences \citep{Addis2007-et}.
In contrast, semantic memory refers to the memory of general knowledge, concepts, language, and facts, which can be accessed without relying on specific experiential contexts \citep{semantic-mem}. Explicit memory is primarily associated with higher-level cognitive functions, such as active learning, problem-solving, and language comprehension.
Episodic memory, in particular, is linked to executive functions like self-reflective thinking and future planning \citep{episodic-non-illusion, Schacter2017-vy}, while semantic memory is essential for knowledge-based reasoning, such as language processing and categorization \citep{Binder2011-cx}.
The hippocampus and medial temporal lobe are responsible for these processes \citep{SQUIRE2004171}.

\textbf{Implicit Memory} Implicit memory operates mainly by influencing performance and behavior through unconscious learning \citep{implicit-mem}.
It is typically divided into procedural memory, classical conditioning, and priming.
Procedural memory mainly involves motor skills and habits, such as riding a bicycle or playing a musical instrument, which are performed automatically without needing explicit recall \citep{motor}.
These skills are developed through repeated practice and learning, allowing them to be executed effortlessly \citep{implicit-dissociation}.
Classical conditioning forms stimulus-stimulus associations, while habit learning forms stimulus-response associations through extensive repetition \citep{classical-condition, habits-rituals}.
Implicit memory is a key factor in cognitive and behavioral abilities such as habituation, automated skill performance, and emotional learning.
This type of memory is closely connected to brain structures like the amygdala, cerebellum, and basal ganglia \citep{implicit-circuit}.

\subsection{Formation Mechanism}
\label{subsec:formation-mechanism}

\textbf{Explicit Memory} The formation of explicit memory occurs through strong synaptic strengthening in the main neural circuits of the hippocampus and medial temporal lobe when new events are encountered \citep{hm-hippocampus, medial-temporal-memory}.
The hippocampus encodes events through sparse representations that index distributed cortical patterns of experience \citep{hipo.20350, Goode2020}.
Specifically, the entorhinal cortex (EC) serves as the gateway between the neocortex and hippocampus, relaying experience into the hippocampal trisynaptic pathway (EC $\rightarrow$ dentate gyrus $\rightarrow$ CA3 $\rightarrow$ CA1). This pathway facilitates memory encoding, where each stage exhibits distinct synaptic plasticity \citep{hippocampal-pathway}.

First, in the dentate gyrus (DG), neural signals are sparsely encoded \citep{sparse-coding}, selectively activating a small subset of neurons and thereby strongly inducing long-term potentiation (LTP). If the neurons were densely coded, a large number of neurons would be redundantly activated, leading to overlap and interference between distinct concepts and episodes. However, sparse coding ensures that even similar information is clearly separated (pattern separation), allowing distinct and non-overlapping memory representations \citep{pattern-separation, leutgeb2007pattern}.

Next, the CA3 region binds these sparsely active neurons through its recurrent collateral circuit, forming an autoassociative network.
Because NMDA receptor-dependent LTP can be induced by only a few stimulations, CA3 rapidly stores the co-active elements of an episode as a single bound pattern that can later be recovered from a partial cue \citep{ca3-nmda, ca3-circuit}.

Finally, the CA1 region serves as the output stage of the hippocampus \citep{ca1-hippo-output}, integrating information received from CA3 and transmitting it to the cerebral cortex and other brain areas.
By regulating the hippocampal output pathways, CA1 contributes to the comprehensive management of memory, facilitating the transition of temporary hippocampal memories into long-lasting neocortical representations \citep{ca1-ltp-ltd}.

\textbf{Implicit Memory} Implicit memory is primarily formed in subcortical structures such as the basal ganglia, cerebellum, and amygdala \citep{implicit-basal-gang-circuit, Conway2020-vv}. Among these, the basal ganglia are crucial in decision-making for behavior through the cortico-basal ganglia-thalamocortical loop \citep{implicit-basal}, which repeatedly receives dense neuronal inputs from the cerebral cortex. For example, Medium Spiny Neurons (MSNs) in the striatum of the basal ganglia express D1 (direct pathway) and D2 (indirect pathway) dopamine receptors and respond sensitively to reward prediction and dopamine signals \citep{basal-automatic, implicit-dopamine}.

The main stages of learning occur as follows:
\begin{enumerate}[topsep=0pt,itemsep=1ex,partopsep=1ex,parsep=0ex,leftmargin=1.5em]
    \item \textbf{Prediction}: Anticipating a reward based on performing a specific behavior in the current context.
    \item \textbf{Action}: When the behavior is performed, signals from the cortex are densely transmitted to the striatal MSNs.
    \item \textbf{Outcome Evaluation}: Assessing whether the behavior resulted in a reward, failure, or punishment.
    \item \textbf{Synaptic Changes}: Strengthening the direct pathway if the reward is greater than expected, strengthening the indirect pathway otherwise.
\end{enumerate}

Since dopamine-driven synaptic strengthening or weakening happens after behavior, a one-time result typically does not cause significant synaptic change \citep{prediction-and-reward}. Repeated trials are necessary to solidify the connection between a specific situation and the corresponding behavior, creating a stimulus-action mapping. Once a behavior becomes part of procedural memory or habit, it is performed or inhibited automatically when the same situation occurs, demonstrating implicit learning. Overall, implicit memory develops gradually over time, requiring numerous repetitions and trial-and-error \citep{dopamine-computation}.

\textbf{Comparison} Explicit memory activates a selective set of neurons (sparse coding) to create clear memories. In contrast, implicit memory relies on behavior and outcomes, gradually recruiting many neurons (dense coding) that operate simultaneously. Explicit memory forms rapidly and powerfully, whereas implicit memory develops through long-term repetition, marking a key difference between the two types of memory \citep{declarative-and-nondeclarative, comparison-memories}.

\subsection{Retrieval Mechanism}

\textbf{Explicit Memory} Explicit memory retrieval centers on the hippocampal-cortical loop, where contextual cues are essential \citep{hippo-neo-cotex, hippo-retrieval, KUMARAN2016512}.
For example, when presented with the word ``summer'', this cue is transmitted through the entorhinal cortex (EC) to activate the CA3 region.
The CA3 region performs pattern completion through its recurrent circuitry, which can reconstruct an entire stored memory pattern even when only a few neurons are activated \citep{ca3-nmda, pattern-comp-sep}.
For instance, the word ``summer'' might activate neurons associated with sensory elements like the beach, sand, the sound of waves, and the smell of the breeze.

This reconstructed memory pattern is then relayed back to the broader cortex \citep{cortical-hippo, cortical-hippo2}. Passing through the CA1 region, the pattern interacts bidirectionally with various areas of the cerebral cortex, such as the prefrontal cortex, sensory cortex, and association cortex, synchronizing different elements of the memory \citep{memory-activatioin-consolidation, Mattar2018}.

In summary, the process of retrieving explicit memory can be outlined as follows:

\begin{enumerate}[topsep=0pt,itemsep=1ex,partopsep=1ex,parsep=0ex,leftmargin=1.5em]
    \item A contextual cue or intention initiates the hippocampal-cortical loop.
    \item Signals from the EC activate specific neurons in the CA3 region.
    \item The CA3 region's recurrent circuitry reconstructs the entire memory pattern based on the activated neurons.
    \item The reconstructed pattern is relayed through the CA1 region to the cerebral cortex, synchronizing the elements of the memory.
    \item Top-down attention from the prefrontal cortex modulates the retrieval process, enabling coherent and integrated memory to emerge consciously.
\end{enumerate}

This process allows rich and multi-layered memories to be replayed from a single cue or piece of information.
Through this process, the hippocampus reactivates the distributed neural patterns of a past episode, effectively recreating the entire context of the original experience.

\textbf{Implicit Memory} Implicit memory retrieval happens automatically \citep{habits-rituals, basal-automatic}. When a familiar stimulus appears, stored stimulus-response patterns in the striatum are activated via MSNs \citep{role-basal-gang-habit}. For example, if you suddenly need to stop while driving, your foot instinctively presses the brake. This occurs because cortical signals reach the striatum, exciting MSNs previously associated with that action \citep{basal-gang-msn}. The MSNs then facilitate action selection through either the direct (D1) or indirect (D2) pathway. The direct pathway enables the action, while the indirect pathway suppresses certain actions \citep{dopamine-modulation}. In short, implicit memory retrieval reactivates learned behaviors automatically in response to familiar cues.

\textbf{Comparison} Explicit and implicit memory retrieval operate through distinct mechanisms tailored to their unique functions.
In explicit memory retrieval, sparse coding plays a key role in pattern completion.
A small number of neurons, activated by a cue, trigger the reconstruction of the entire memory through highly selective connections.
This process allows the memory system to retrieve detailed and multi-faceted experiences by filling in missing details and linking related sensory and contextual elements into a coherent whole.
Implicit memory, on the other hand, operates via learned stimulus-response patterns.
When a familiar stimulus is encountered, the system activates the corresponding response pattern automatically, allowing previously learned behaviors or routines to be executed efficiently.
Unlike explicit memory, implicit retrieval does not reconstruct a broader memory context but instead focuses on directly accessing and applying specific learned associations.

\section{Nature of Learning in LLMs}
\label{sec:nature-of-learning-llms}

This section examines the learning and reasoning processes of LLMs, focusing on how their low-level mechanisms relate to explicit and implicit memory systems.
In doing so, it reveals key distinctions from hippocampal explicit memory and indicates that the core characteristics of LLMs more closely resemble implicit learning and habituation.

\textbf{Gradual Learning} Explicit memory can encode new episodes vividly with just one exposure when strong stimuli are present or attention is engaged. In contrast, implicit memory relies on gradually accumulating associations between stimuli, actions, and rewards (or punishments).
For instance, repeated input projection strengthens or weakens synapses incrementally, guided by dopamine signals reflecting reward prediction errors.
Similarly, LLM training follows a comparable pattern.
Model parameters do not change drastically with one or two examples.
Instead, the model adjusts its weights cumulatively by repeatedly observing numerous examples within a vast corpus, aiming to reduce errors (loss).
This gradual and repetitive learning mirrors the characteristics of implicit memory formation.

\textbf{Dense and Distributed Coding} Explicit memory is characterized by sparse coding to ensure pattern separation. In contrast, subcortical circuits receive dense signals from the cortex.
LLMs exhibit a similar structure, with their network broadly interconnected.
Their activations for specific contexts are not confined to isolated nodes but occur across a wide range of parameters.
These features of distributed coding and large-scale parallel activations are closely aligned with the implicit processing of the basal ganglia.

\textbf{Error-Driven Learning} Implicit learning involves dopamine neurons strengthening or weakening synapses based on reward prediction errors.
In contrast, learning in explicit memory relies on the associations of events or pieces of information, rather than on an error-based mechanism.
The hippocampus rapidly encodes new events, combining elements presented together in time and space into a single episode.
For LLMs, the loss function works in a similar way to dopamine signals, adjusting model parameters through backpropagation based on the difference between the actual output and the ground truth across large datasets.
Thus, the error minimization mechanism based on prediction errors is a key shared feature of implicit memory and LLMs, distinguishing them from hippocampal learning.
This distinction suggests that the underlying learning mechanism of LLMs is closely aligned with implicit learning.

\textbf{Automatic Action} Explicit memory retrieval through the hippocampus can reconstruct past episodes richly from just one contextual cue.
In contrast, implicit memory retrieval in the striatum triggers automatic procedural responses without reconstructing any past episodes.
In the case of LLMs, when specific prompts (contexts) are provided, the process does not involve reconstructing an entire episode like the hippocampus.
Instead, LLMs follow a procedure where their learned parameters automatically determine the optimal next word.
This resembles how humans unconsciously execute habits or conditioned responses based on implicit memory, suggesting that LLM retrieval mechanisms are based on the stimulus-response mapping of the basal ganglia.

\textbf{Fundamental Similarities} In summary, the low-level learning mechanism of LLMs is closer to implicit memory systems, specifically habituation and procedural learning, than to explicit memory processes. This perspective provides valuable insights into understanding LLMs as systems that require gradual and repetitive training, and exhibit the automatic retrieval characteristics of human implicit learning.

\section{Functional Specificity of Memory}
\label{sec:functional-specificity-mem}

Explicit and implicit memory serve distinct cognitive roles and are not easily interchangeable. This section explores the unique cognitive abilities supported by each memory system.
By examining these differences, I indirectly reveal the cognitive functions that LLMs, which parallel implicit memory, are capable of possessing and those they cannot.

\textbf{Pattern Recognition} Pattern recognition is the ability of humans to identify specific rules or consistencies in a complex environment.
For instance, people can quickly analyze facial features like eyes, nose, and mouth to identify a specific person.
In language comprehension, even when spelling is incorrect or sentences are incomplete, humans can capture the meaning by recognizing patterns in words and context.

Research on statistical learning demonstrates that human pattern recognition and perception are deeply rooted in the implicit memory system.
\citet{Fiser2002-aslin, Fiser2002-kt} showed that infants have the ability to automatically learn statistical patterns of co-occurrences and conditional probabilities between visual objects.
\citet{Turk-Browne2005-tu} revealed that visual statistical learning is automatic and implicit, despite being gated by selective attention.
Additionally, \citet{BATTERINK201731} used EEG measurements to track brain activity related to implicit learning while participants detected patterns in word sequences.
Ultimately, pattern recognition is an automated information processing function based on implicit memory, which is consistently supported by historical theory and modern experimental evidence.

\textbf{Language Ability} The interaction between explicit memory and implicit memory in language processing has been demonstrated through various studies, emphasizing that fluent communication heavily relies on implicit memory \citep{implicit-word-completion}. Ullman's Declarative/Procedural model \citep{Ullman2004-hj} posits that the mental lexicon, which stores word-specific knowledge, depends on explicit memory, while implicit memory plays a key role in the automatic application of grammatical rules.
Research on patients with hippocampal damage \citep{Vargha-Khadem1997-jj, Schmolck2002-ul--hm} showed that even when episodic memory is impaired, semantic memory remains intact, enabling normal language abilities.
This suggests that language fluency operates independently of explicit memory.

In contrast, studies revealed that dysfunctions in the basal ganglia disrupt grammatical processing and automated language use, highlighting the critical contribution of implicit memory in procedural language functions \citep{Lieberman1992-lv, Ullman1997-dy}.
Finally, \citet{Booth2007-fr} showed that the basal ganglia and cerebellum interact with brain regions responsible for phonological processing, supporting the refinement and automation of linguistic processes.
This underscores the essential role of implicit memory in natural and flexible communication.
Therefore, language ability can be understood as relying on a dual memory system, where explicit memory processes new information during the initial stages of learning, and repeated use transitions it into implicit memory for automation.

\textbf{Logical Reasoning} Mathematical and logical reasoning rely heavily on semantic memory, a subset of explicit memory, as evidenced by various neuroscience studies.
\citet{Friedrich2013-ps} found that the interaction between the prefrontal cortex and hippocampal systems is essential for processing the semantics of mathematical logic.
\citet{Menon2016-nr} further explained that declarative memory is an integral part of forming associative memories, which allows for generalization beyond superficial problem attributes.
Additionally, research by \citet{Evans2024-pr} showed that declarative memory is a key predictor of elementary school students' mathematical skills, whereas procedural memory has little influence.
Although \citet{FAYOL2012392} observed that simple operations like addition and subtraction can be performed through procedural memory, the converging evidence suggests that complex mathematical and logical reasoning critically depends on explicit memory.

\textbf{Executive Function} Executive functions are key components of higher-order cognitive abilities in humans. They encompass various subdomains that help achieve goals, solve problems, and adapt to changing environments. In particular, cognitive flexibility, planning, decision-making, and task-switching are critical skills that enable individuals to adapt to changes, achieve goals systematically, make optimal decisions, and efficiently transition between tasks.
Explicit memory is a crucial factor in executive functions through the retrieval of past experiences, which supports complex cognitive processes such as planning and decision-making.

\citet{Klein2002-ce} explained that explicit memory provides contextually relevant information to decision-making systems, enabling more accurate judgments.
This role of memory is also demonstrated in multitasking studies by \citet{Burgess2000-hq}, who found that retrospective memory supports prospective memory and planning abilities, while damage to related brain regions can impair these functions.
\citet{Hassabis2007-vo} emphasized that the scene construction process of episodic memory is crucial for future-oriented thinking and goal-directed actions.
\citet{Whittington-cell} computationally demonstrated that the hippocampal formation organizes knowledge into structured relational maps.

On the other hand, deficits in explicit memory are closely linked to impaired executive functions.
For example, \citet{Johns2012-om} reported that individuals with Down syndrome who performed poorly on explicit memory tasks showed weaknesses in executive functions such as working memory and cognitive flexibility.
Lastly, \citet{Pedraza-24} demonstrated that implicit statistical learning can compete with executive functions and even negatively impact declarative learning processes, such as goal-directed behaviors.
In summary, explicit memory provides the foundation for executive functions by enabling the retrieval and manipulation of information.
It is essential for higher-order processes like planning, judgment, and future-oriented thinking.

\textbf{Metacognition} Explicit processes are central to metacognition and reflection, enabling individuals to monitor and regulate their cognitive behaviors.
The feeling of knowing (FOK) exemplifies this, as explored by \citet{Irak2019-lc}, who linked FOK judgments to specific neural components, emphasizing the role of explicit mechanisms in predicting memory retrieval. 
\citet{NELSON1990125} similarly highlighted that FOK relies on explicit control and monitoring processes to resolve the paradox of predicting recognition of inaccessible items.
\citet{Fleming2012-wm} further tied metacognitive accuracy to the prefrontal cortex, showing that reflective judgments depend on explicit representations.
This aligns with \citet{Klein2010-pb}'s view of the self as a system of interrelated explicit processes, crucial for integrating episodic memory and semantic knowledge.
In nonhuman primates, \citet{Hampton2020-mf} provided evidence of explicit memory systems supporting metacognition, reinforcing their evolutionary significance.
These studies illustrate that explicit mechanisms are essential for reflective thought, enabling self-awareness and adaptive decision-making.

\textbf{Mental Simulation} Episodic memory, as a constructive and flexible system, plays a critical role in mental simulation by enabling the recombination of past experiences to envision future scenarios \citep{episodic-non-illusion}.
This capacity for ``mental time travel'' is uniquely human and vital for adaptive planning \citep{Suddendorf2007-dh}.
Episodic memory and simulation share the default mode network, which supports scene construction and autobiographical thought, dividing into distinct subsystems handling memory retrieval and reflective social cognition \citep{Hassabis2007-vo, Andrews-Hanna2014-zs}.
Moreover, while episodic and semantic details interact dynamically, episodic simulation often demands greater cognitive control to construct vivid hypothetical scenarios \citep{Benoit2015-rc, Devitt2017-ew}.
Together, these findings highlight how episodic memory underpins the mental simulations that enable humans to anticipate and navigate the future.

In conclusion, cognitive functions are closely linked to specific memory systems, and understanding these dynamics allows us to better identify the strengths and weaknesses of LLMs. Due to their exceptional performance in implicit memory-based processes, LLMs excel at recognizing patterns, generating fluent language, and leveraging statistical associations within large datasets. However, their various limitations become apparent in tasks requiring explicit memory, which will be discussed in detail in the next section.

\section{Need for Explicit Memory}
\label{sec:need-for-explicit-memory}

The previous section examined how explicit and implicit memory systems influence distinct cognitive functions. This section addresses the key limitations that LLMs must overcome to evolve into AGI.
It is notable that these limitations closely align with the inherent constraints of implicit memory, which lacks mechanisms for rapid and context-rich learning.
Explicit memory, in contrast, offers a potential pathway to overcoming these challenges by enabling higher-order cognitive abilities of humans.
Refer to \cref{sec:evidence-absence-exmem} for actual examples showing these weaknesses.

\subsection{Higher-Order Learning}
\label{subsec:higher-order-learning}

Humans utilize a wider range of higher-order learning mechanisms based on explicit memory.
First, humans are capable of \textbf{dynamic learning}, continuously acquiring new knowledge over time.
In contrast, the learning process in LLMs is strictly confined to the training stage, which limits their potential to function as real-time learners.
\textbf{One-shot learning} is especially crucial for immediate adaptation of agents in settings such as narrative memory, personalized contexts, or unpredictable situations. 
This instant learning is also enabled by explicit memory, through substantial synaptic weight adjustments after just a single experience.

Furthermore, \textbf{episodic learning}, which combines knowledge with its source and spatiotemporal context, forms the foundation of the ``feeling of knowing'' and metacognition.
This type of learning grants individuals their unique self-narratives.
Humans can also learn through semantic memory, which allows the understanding of abstract relationships found in books without requiring direct experience or links to rewards.
This capacity for \textbf{abstract learning} is a distinctive ability that sets humans apart from other primates.

Finally, human explicit memory systems are built on associations between knowledge structures.
When a new concept is introduced, it is not treated as an isolated sample.
Instead, it integrates into existing semantics by forming associations with them, enabling \textbf{continual learning} by updating or expanding knowledge in a natural way.
This higher-order learning is essential for bridging the gap between LLMs and AGI and strongly supports the need for explicit memory.

\subsection{Metacognition}

\textbf{Hallucination} has been a well-known and persistent issue in LLMs since their early stages.
In essence, hallucination can be attributed to a lack of \textbf{metacognition}.
Metacognition entails the ability to distinguish between what one knows and does not know, as well as to keep track of the source and context of one's knowledge.
Since LLMs lack episodic metadata about their knowledge and learn by forming probabilistic distributions based on data, the occurrence of hallucinations is an expected byproduct. 

The problem of \textbf{consistency}, where LLMs provide different answers to the same question under subtle noise, is also inevitable due to the absence of metacognition and any grounding for the knowledge.
In contrast, humans can store their actions as memories in specific situations, reflect on the outcomes later, and engage in \textbf{mental simulations} or \textbf{reflections}.
This ability relies on episodic memory, which is essential for cognitive abilities related to metacognition.

\subsection{Logical Reasoning}

As mentioned earlier, simple arithmetic is often automatically mapped into human procedural memory through repeated practice.
However, when \textbf{arithmetic problems} become more complex and involve longer numbers, humans can solve them given enough time, while LLMs tend to struggle.
Logical reasoning, such as \textbf{mathematical reasoning, causal reasoning, or deductive reasoning}, heavily relies on concepts stored in semantic memory and the associations between them.
The difficulty of logical reasoning lies in the fact that the core logical concepts and their relationships are so abstract that the possible variations of corresponding low-level representations can be practically infinite.
For example, the principle of multiplication is simple, but the specific expressions representing multiplication equations are infinite.
Such representational diversity makes statistical learning difficult.
Solving such high-level logical problems using only statistical habituation, without relying on semantic memory, would be nearly impossible.

\subsection{Executive Function}

Executive functions encompass a variety of sub-cognitive functions. Among these, regulatory abilities like \textbf{inhibition} and \textbf{task-switching} are not prominently displayed in current LLMs, as they are not necessary without episodic memory.
\textbf{Planning}, while long considered a weakness of LLMs, has seen significant improvement with repeated efforts and the development of models specialized in reasoning.
Executive functions heavily rely on short-term storage, often referred to as working memory.
Within the context window that corresponds to working memory, LLMs demonstrate strong reasoning capabilities, allowing for planning within a limited scope.
However, to reach human-level capabilities, such as writing a 1,000-page book or completing a year-long project broken down into daily tasks, LLMs must maintain coherence beyond their context limits, requiring episodic memory.
If such advancements are realized, the need for regulation and task-switching will become evident, as they will allow systematic goal-setting and the ability to filter out unnecessary tasks while carrying out the work.

\section{Computational Requirements for Artificial Explicit Memory System}
\label{sec:computational-requirements}

In this section, I define an explicit memory system that can artificially replicate the aforementioned functions and specify the computational properties required for such a memory system. The memory system is treated as a module that receives a dense embedding from upstream representational processes. The system is defined as a function $f_{\mathrm{memory}}$:
\begin{equation}
  \begin{aligned}
      f_{\mathrm{memory}} &: I \rightarrow O \\
      I &= (E, M) \\
      O &= (\Delta M, Y).
  \end{aligned}
\end{equation}
The input consists of the current dense embedding $E \in \mathbb{R}^d$ and the current memory state $M$. The output comprises the memory update $\Delta M$ and the retrieved dense embedding $Y \in \mathbb{R}^d$. The internal structure of $M$ is specified incrementally through the following requirements.

\textbf{Sparse Indexing} The memory system generates a sparse index from the dense input, where most elements are zero:
\begin{equation}
    \begin{aligned}
        S &= \mathrm{sparsify}(E) \in \mathbb{R}^n, \\
        \|S\|_0 &= \sum_{i=1}^{n} \mathbf{1}[S_i \neq 0] \ll n.
    \end{aligned}
\end{equation}
Let $F_S = \{i \mid S_i \neq 0\}$ denote the set of activated dimensions. The memory state $M$ includes a key-value mapping $P \in \mathbb{R}^{n \times d}$, where each row $P_i \in \mathbb{R}^d$ represents the dense embedding referenced by sparse index dimension $i$. The mapping $P$ binds the sparse index to its dense source:
\begin{equation}
    P^\top S \approx E.
\end{equation}
This binding ensures that each active sparse dimension functions as a pointer to its corresponding dense representation.

\textbf{Error-Independent Update} Memory updates should not be driven by error minimization; instead, $\Delta M$ should mainly be computed as a function of $E$ and $M$, with no direct dependence on prediction $p$ or prediction error $e$:
\begin{equation}
    \nabla_{p,e} \Delta M = \mathbf{0}.
\end{equation}
This does not mean memory must be entirely decoupled from prediction. Predictions or perceived errors may still influence memory. What the constraint excludes is the case where prediction error dominantly drives the update via error-driven learning systems such as gradient descent.

\textbf{Associative Construction} The memory state $M$ further includes an associative matrix $A \in \mathbb{R}^{n \times n}$ that captures relationships among sparse indices. Each piece of information is not stored independently but is interconnected with the existing memory. For a given sparse index $S$, the update $\Delta A$ strengthens the connections among all activated dimensions, following the principle of ``fire together, wire together'':
\begin{equation}
    \Delta A_{i,j} > 0, \quad \forall i, j \in F_S.
\end{equation}

\textbf{Pattern Separation} Sparse indexing alone does not guarantee separation. $\mathrm{sparsify}$ must additionally satisfy a non-expansion property such that sparse codes are less similar than their dense inputs. Formally, for an appropriately normalized semantic similarity measure $\mathrm{sim}$:
\begin{equation}
    \mathrm{sim}(S_1, S_2) < \mathrm{sim}(E_1, E_2).
\end{equation}
Without this property, two distinct inputs $E_1$ and $E_2$ could collapse to overlapping sparse codes, becoming indistinguishable to the associative matrix $A$ and undermining both pattern completion and key-value lookup.

\textbf{Pattern Completion} An explicit memory system can restore information from partial input through pattern completion, which proceeds in two stages. Given a partial input $E^{\mathrm{partial}}$, the system first generates the corresponding sparse index, then performs index completion via $A$ and dense reconstruction via $P$:
\begin{equation}
    \begin{aligned}
        S^{\mathrm{partial}} &= \mathrm{sparsify}(E^{\mathrm{partial}}), \\
        S^{\mathrm{retrieved}} &= \sigma(A \, S^{\mathrm{partial}}), \quad F_{S^{\mathrm{partial}}} \subset F_{S^{\mathrm{retrieved}}}, \\
        Y &= P^\top S^{\mathrm{retrieved}} \in \mathbb{R}^d.
    \end{aligned}
\end{equation}
Here, $\sigma$ denotes any selective readout that preserves sparsity (e.g., top-$k$, thresholding, or winner-take-all). If $E^{\mathrm{learned}}$ was previously stored and the partial sparse code is a subset of its full code, $F_{S^{\mathrm{partial}}} \subset F_{S^{\mathrm{learned}}}$, then $Y \approx E^{\mathrm{learned}}$.

\textbf{Dynamicity} The memory state must evolve over time, with changes $\Delta M_t$ at time $t$ depending on the input $E_t$ and the current memory state $M_t = (A_t, P_t)$:
\begin{equation}
    \begin{aligned}
        I_t &= (E_t, M_t), \\
        O_t &= (\Delta M_t, Y_t), \\
        M_{t+1} &= M_t + \Delta M_t = (A_t + \Delta A_t,\; P_t + \Delta P_t).
    \end{aligned}
\end{equation}

\textbf{High and Instant Plasticity} Explicit memory must enable immediate storage of information after a single experience. Once $E_t$ is learned at time $t$, the system must have the capacity to retrieve it via pattern completion at the immediate following timestep $t+1$ using only partial input $E_t^{\mathrm{partial}}$:
\begin{equation}
    \begin{aligned}
        f_{\mathrm{memory}}(E_t^{\mathrm{partial}}, M_{t+1}) &= (\Delta M_{t+1}, Y_{t+1}), \\
        Y_{t+1} &\approx E_t.
    \end{aligned}
\end{equation}

\textbf{Adaptive Forgetting} Memory capacity is necessarily bounded, and the encoding of new patterns could come at the cost of existing connections. When a new sparse pattern $S$ is encoded, connections between active and inactive dimensions undergo heterosynaptic weakening:
\begin{equation}
    i \in F_S, \; j \notin F_S \implies \Delta A_{i,j} \leq 0.
\end{equation}
Only connections involving currently active dimensions undergo plasticity, while connections between two inactive dimensions remain undisturbed.

This section outlines the eight computational requirements necessary for an explicit memory system to enable artificial neural networks to acquire higher-order cognitive functions. These requirements identify the most essential conditions rather than provide a complete biological replica; thus, simplifications such as the use of a unified bidirectional $P$ or the omission of cortical consolidation remain areas for future refinement. I review the findings of recent studies through the lens of these computational requirements in \cref{sec:recent-progress}.

\section{Alternative Views}
\label{sec:alternative-views}

As highlighted throughout this paper, there is a fundamental gap between current LLMs and AGI. Existing approaches confined to implicit learning mechanisms alone cannot bridge this divide. I argue that integrating an explicit memory system is essential to overcome this limitation, though this position naturally invites alternative perspectives.

The most direct counterargument is that current models have already crossed the threshold of AGI. Microsoft researchers presented their view of GPT-4 as an early, incomplete AGI \citep{ms-gpt4-agi}, while a researcher at OpenAI claimed AGI has been achieved with models like o1. However, these are minority views, and the dominant opinion is that AGI or Human-Level AI has not been realized.

Beyond the definition of AGI, another viewpoint holds that explicit memory is not necessarily required to achieve AGI.
However, since the discussion on the necessity of explicit memory is still in its early stages, counterarguments against it are not yet widely articulated.
Indeed, memory is generally considered an essential component of AGI \citep{bio-underping, generative-ai-vs-agi, how-far-agi-llm}. The debate is more likely to focus on alternative ways to integrate memory into AI rather than questioning its necessity.

Complementary Learning Systems theory \citep{hippo-neo-cotex, KUMARAN2016512} shares my premise that a hippocampus-like module is needed but treats LLMs as analogues of neocortical semantic memory. However, as \cref{sec:evidence-absence-exmem} shows, LLMs fail to produce the rule-respecting, flexibly retrievable knowledge that defines semantic memory, reverting to rigid stimulus–response behavior. Thus, I maintain that LLMs operate fundamentally as implicit systems. \cref{subsec:difference-cls} provides an extended discussion.

\citet{ai-native} proposed a solution using LLMs as processors while integrating AI-native long-term memory in a systematic way to achieve AGI.
This approach first builds a natural language memory, which is then compressed into a neural network-based personal model to enable continuous learning and personalized reasoning. Through this process, AI is expected to develop personalized intelligence and adaptive capabilities, bringing it closer to AGI.


\section{Discussion}
\label{sec:discussion}

\textbf{Current Success} If explicit memory is as essential, one might expect current LLMs to be more limited than they are, given that they already handle tasks well beyond the reach of human implicit memory. I attribute this to two primary factors. The first is scale. A key constraint on human implicit memory is the experience and time it requires, yet LLMs are exposed to more text than a person reads in a lifetime, pushing implicit-style learning to a biologically unattainable degree. The second factor is optimization. Although implicit learning is error-driven, the brain cannot optimize via per-weight gradients as artificial networks do. Storing a gradient for every weight is an overhead no organism could sustain, and exploiting it yields clear gains in certain domains. In short, the success of LLMs does not contradict their alignment with implicit memory; humans are simply bound by biological limits far tighter than the theoretical ceiling of the implicit system, and have never fully realized its potential.
Refer to \cref{subsec:potential-limit-implicit} for further discussion.

\textbf{Testability} The central claim of this paper rests on the premise that current LLMs are fundamentally implicit memory systems and therefore lack explicit memory. This hypothesis must ultimately be confirmed or refuted empirically. Showing that a model does not operate explicitly is relatively straightforward. As shown in \cref{sec:evidence-absence-exmem}, when a model fails on problems requiring the exact same knowledge it successfully applies in closely related cases, it indicates a lack of explicit processing. Proving the converse, however, is significantly more challenging. Because LLMs serve an enormous user base, their failures are rapidly collected and integrated back into training data by providers. Consequently, even when such problems are eventually resolved, the improvement cannot prove that the task is now solved explicitly rather than through expanded statistical coverage.

The most reliable behavioral evidence for explicit learning therefore requires deliberate omission from the training data together with a test of adaptation to the omitted task. For instance, if a model trained without any multiplication problem sets could perform compound multiplication after learning only addition and the definition of multiplication, this would be strong evidence of explicit memory. An indirect signal comes from tasks that require tracking a rapidly changing state over a long horizon, such as complex entity tracking, which lies beyond the reach of implicit memory. It is essential that the state not be written explicitly into the tokens; otherwise, the task reduces to a chain of automatic rule applications and no longer probes explicit processing.

\textbf{Further Requirements} Reaching AGI necessitates not only the development of the explicit memory system itself but also broader theoretical considerations such as determining the precise nature of the input information (e.g., contextual episodes, self-actions, long-term temporal dependencies) and how explicit memory can interact with other forms of memory. Furthermore, because the proposed memory system functions by indexing representations produced by upstream computation, its effectiveness fundamentally depends on whether the LLM's internal reasoning itself operates in an explicit manner. This remains an open question that warrants dedicated investigation. See \cref{subsec:actional-subtopics}.

\section{Conclusion}
\label{sec:conclusion}

The path to AGI requires addressing significant limitations in current LLMs. While LLMs excel in tasks that involve implicit memory, such as pattern recognition and statistical learning, they lack explicit memory capabilities.
This limits their ability to perform dynamic learning, long-term reasoning, and metacognitive tasks essential for AGI.

Explicit memory is critical for storing and retrieving episodic and semantic information, enabling models to update knowledge over time, reason logically, and simulate hypothetical scenarios.
These capabilities allow LLMs to overcome challenges like hallucination, inconsistency, and deficiencies in long-term planning.
Key features include sparse coding, pattern completion, and dynamic updating mechanisms, which mirror human memory processes.

To achieve AGI, research can significantly benefit from prioritizing the integration of explicit memory into LLMs and exploring its interaction with other memory systems.
This shift will empower LLMs to perform complex, coherent tasks and adapt in real-time, moving closer to the versatility and depth of human cognition.
I hope that the theoretical framework will guide the evolution of robust explicit memory systems and contribute to the broader NeuroAI agenda of leveraging validated biological principles to advance artificial intelligence \citep{Hassabis2017-ul, Zador2023}.

\bibliography{main}

@article{cortical-hippo2,
	title        = {Declarative memory consolidation in humans: A prospective functional magnetic resonance imaging study},
	author       = {A. Takashima  and K. M. Petersson  and F. Rutters  and I. Tendolkar  and O. Jensen  and M. J. Zwarts  and B. L. McNaughton  and G. Fernández},
	year         = {2006},
	journal      = {Proceedings of the National Academy of Sciences},
	volume       = {103},
	number       = {3},
	pages        = {756--761},
	doi          = {10.1073/pnas.0507774103},
	url          = {https://www.pnas.org/doi/abs/10.1073/pnas.0507774103},
	eprint       = {https://www.pnas.org/doi/pdf/10.1073/pnas.0507774103}
}

@article{bigbench,
	title        = {Beyond the Imitation Game: Quantifying and extrapolating the capabilities of language models},
	author       = {Aarohi Srivastava and Abhinav Rastogi and Abhishek Rao and Abu Awal Md Shoeb and Abubakar Abid and Adam Fisch and Adam R. Brown and Adam Santoro and Aditya Gupta and Adri{\`a} Garriga-Alonso and Agnieszka Kluska and Aitor Lewkowycz and Akshat Agarwal and Alethea Power and Alex Ray and Alex Warstadt and Alexander W. Kocurek and Ali Safaya and Ali Tazarv and others},
	year         = {2023},
	journal      = {Transactions on Machine Learning Research},
	issn         = {2835-8856},
	url          = {https://openreview.net/forum?id=uyTL5Bvosj},
	note         = {Featured Certification}
}

@article{Addis2007-et,
	title        = {Remembering the past and imagining the future: Common and distinct neural substrates during event construction and elaboration},
	author       = {Addis, Donna Rose and Wong, Alana T and Schacter, Daniel L},
	year         = {2007},
	journal      = {Neuropsychologia},
	publisher    = {Elsevier Science},
	address      = {Netherlands},
	volume       = {45},
	number       = {7},
	pages        = {1363--1377}
}

@incollection{working-mem,
	title        = {Working Memory},
	author       = {Alan D. Baddeley and Graham Hitch},
	year         = {1974},
	publisher    = {Academic Press},
	series       = {Psychology of Learning and Motivation},
	volume       = {8},
	pages        = {47--89},
	doi          = {https://doi.org/10.1016/S0079-7421(08)60452-1},
	issn         = {0079-7421},
	url          = {https://www.sciencedirect.com/science/article/pii/S0079742108604521},
	editor       = {Gordon H. Bower}
}

@article{implicit-basal-gang-circuit,
	title        = {Parallel organization of functionally segregated circuits linking basal ganglia and cortex},
	author       = {Alexander, G E and DeLong, M R and Strick, P L},
	year         = {1986},
	journal      = {Annu. Rev. Neurosci.},
	publisher    = {Annual Reviews},
	volume       = {9},
	number       = {1},
	pages        = {357--381},
	language     = {en}
}

@misc{qwen25,
	title        = {Qwen2.5 Technical Report},
	author       = {An Yang and Baosong Yang and Beichen Zhang and Binyuan Hui and Bo Zheng and Bowen Yu and Chengyuan Li and Dayiheng Liu and Fei Huang and Haoran Wei and Huan Lin and Jian Yang and Jianhong Tu and Jianwei Zhang and Jianxin Yang and Jiaxi Yang and Jingren Zhou and Junyang Lin and Kai Dang and others},
	year         = {2025},
	url          = {https://arxiv.org/abs/2412.15115},
	eprint       = {2412.15115},
	archiveprefix = {arXiv},
	primaryclass = {cs.CL}
}

@article{hippocampal-pathway,
	title        = {Lamellar organization of hippocampal excitatory pathways},
	author       = {Andersen, P. and Bliss, T. V. P. and Skrede, K. K.},
	year         = {1971},
	journal      = {Experimental Brain Research},
	volume       = {13},
	number       = {2},
	pages        = {222--238},
	doi          = {10.1007/BF00234087},
	isbn         = {1432-1106},
	url          = {https://doi.org/10.1007/BF00234087},
	date         = {1971/08/01},
	id           = {Andersen1971}
}

@article{acquisition-cog-skill,
	title        = {Acquisition of Cognitive Skill},
	author       = {Anderson, John},
	year         = {1982},
	month        = {07},
	journal      = {Psychological Review},
	volume       = {89},
	pages        = {369--406},
	doi          = {10.1037/0033-295X.89.4.369},
	isbn         = {9781483214467}
}

@article{Andrews-Hanna2014-zs,
	title        = {Contributions of episodic retrieval and mentalizing to autobiographical thought: evidence from functional neuroimaging, resting-state connectivity, and {fMRI} meta-analyses},
	author       = {Andrews-Hanna, Jessica R and Saxe, Rebecca and Yarkoni, Tal},
	year         = {2014},
	month        = may,
	journal      = {Neuroimage},
	publisher    = {Elsevier BV},
	volume       = {91},
	pages        = {324--335},
	language     = {en}
}

@article{titans,
	title        = {Titans: Learning to memorize at test time},
	author       = {Behrouz, Ali and Zhong, Peilin and Mirrokni, Vahab},
	year         = {2024},
	month        = dec,
	journal      = {arXiv [cs.LG]},
	url          = {http://arxiv.org/abs/2501.00663},
	archiveprefix = {arXiv},
	primaryclass = {cs.LG},
	eprint       = {2501.00663}
}

@misc{generative-ai-vs-agi,
	title        = {Generative AI vs. AGI: The Cognitive Strengths and Weaknesses of Modern LLMs},
	author       = {Ben Goertzel},
	year         = {2023},
	url          = {https://arxiv.org/abs/2309.10371},
	eprint       = {2309.10371},
	archiveprefix = {arXiv},
	primaryclass = {cs.AI}
}

@article{Benoit2015-rc,
	title        = {Specifying the core network supporting episodic simulation and episodic memory by activation likelihood estimation},
	author       = {Benoit, Roland G and Schacter, Daniel L},
	year         = {2015},
	month        = aug,
	journal      = {Neuropsychologia},
	publisher    = {Elsevier BV},
	volume       = {75},
	pages        = {450--457},
	language     = {en}
}

@article{Binder2011-cx,
	title        = {The neurobiology of semantic memory},
	author       = {Binder, Jeffrey R and Desai, Rutvik H},
	year         = {2011},
	month        = nov,
	journal      = {Trends Cogn. Sci.},
	publisher    = {Elsevier BV},
	volume       = {15},
	number       = {11},
	pages        = {527--536},
	language     = {en}
}

@article{Booth2007-fr,
	title        = {The role of the basal ganglia and cerebellum in language processing},
	author       = {Booth, James R and Wood, Lydia and Lu, Dong and Houk, James C and Bitan, Tali},
	year         = {2007},
	month        = feb,
	journal      = {Brain Res.},
	publisher    = {Elsevier BV},
	volume       = {1133},
	number       = {1},
	pages        = {136--144},
	language     = {en}
}

@misc{ms-gpt4-agi,
	title        = {Sparks of Artificial General Intelligence: Early experiments with GPT-4},
	author       = {Bubeck, Sébastien and Chandrasekaran, Varun and Eldan, Ronen and Gehrke, Johannes and Horvitz, Eric and Kamar, Ece and Lee, Peter and Lee, Yin Tat and Li, Yuanzhi and Lundberg, Scott and Nori, Harsha and Palangi, Hamid and Ribeiro, Marco Tulio and Zhang, Yi},
	year         = {2023},
	month        = {March},
	url          = {https://www.microsoft.com/en-us/research/publication/sparks-of-artificial-general-intelligence-early-experiments-with-gpt-4/}
}

@article{Burgess2000-hq,
	title        = {The cognitive and neuroanatomical correlates of multitasking},
	author       = {Burgess, P W and Veitch, E and de Lacy Costello, A and Shallice, T},
	year         = {2000},
	journal      = {Neuropsychologia},
	publisher    = {Elsevier BV},
	volume       = {38},
	number       = {6},
	pages        = {848--863},
	language     = {en}
}

@inproceedings{llm-poem,
	title        = {Help me write a Poem - Instruction Tuning as a Vehicle for Collaborative Poetry Writing},
	author       = {Chakrabarty, Tuhin  and Padmakumar, Vishakh  and He, He},
	year         = {2022},
	month        = dec,
	booktitle    = {Proceedings of the 2022 Conference on Empirical Methods in Natural Language Processing},
	publisher    = {Association for Computational Linguistics},
	address      = {Abu Dhabi, United Arab Emirates},
	pages        = {6848--6863},
	doi          = {10.18653/v1/2022.emnlp-main.460},
	url          = {https://aclanthology.org/2022.emnlp-main.460/},
	editor       = {Goldberg, Yoav  and Kozareva, Zornitsa  and Zhang, Yue}
}

@article{classical-condition,
	title        = {Classical conditioning and brain systems: the role of awareness},
	author       = {Clark, R E and Squire, L R},
	year         = {1998},
	month        = apr,
	journal      = {Science},
	publisher    = {American Association for the Advancement of Science (AAAS)},
	volume       = {280},
	number       = {5360},
	pages        = {77--81},
	language     = {en}
}

@article{lake,
	title        = {Building machines that learn and think with people},
	author       = {Collins, Katherine M. and Sucholutsky, Ilia and Bhatt, Umang and Chandra, Kartik and Wong, Lionel and Lee, Mina and Zhang, Cedegao E. and Zhi-Xuan, Tan and Ho, Mark and Mansinghka, Vikash and Weller, Adrian and Tenenbaum, Joshua B. and Griffiths, Thomas L.},
	year         = {2024},
	journal      = {Nature Human Behaviour},
	volume       = {8},
	number       = {10},
	pages        = {1851--1863},
	doi          = {10.1038/s41562-024-01991-9},
	isbn         = {2397-3374},
	url          = {https://doi.org/10.1038/s41562-024-01991-9},
	date         = {2024/10/01},
	id           = {Collins2024}
}

@article{Conway2020-vv,
	title        = {How does the brain learn environmental structure? Ten core principles for understanding the neurocognitive mechanisms of statistical learning},
	author       = {Conway, Christopher M},
	year         = {2020},
	month        = may,
	journal      = {Neurosci. Biobehav. Rev.},
	publisher    = {Elsevier BV},
	volume       = {112},
	pages        = {279--299},
	language     = {en}
}

@article{doi:10.1073/pnas.1403112111,
	title        = {Performance-optimized hierarchical models predict neural responses in higher visual cortex},
	author       = {Daniel L. K. Yamins  and Ha Hong  and Charles F. Cadieu  and Ethan A. Solomon  and Darren Seibert  and James J. DiCarlo},
	year         = {2014},
	journal      = {Proceedings of the National Academy of Sciences},
	volume       = {111},
	number       = {23},
	pages        = {8619--8624},
	doi          = {10.1073/pnas.1403112111},
	url          = {https://www.pnas.org/doi/abs/10.1073/pnas.1403112111},
	eprint       = {https://www.pnas.org/doi/pdf/10.1073/pnas.1403112111}
}

@inproceedings{Das2024-ol,
	title        = {Larimar: Large Language Models with Episodic Memory Control},
	author       = {Das, Payel and Chaudhury, Subhajit and Nelson, Elliot and Melnyk, Igor and Swaminathan, Sarathkrishna and Dai, Sihui and Lozano, Aurelie and Kollias, Georgios and Chenthamarakshan, Vijil and Navratil, Jiri and Dan, Soham and Chen, Pin-Yu},
	year         = {2024},
	month        = jul,
	booktitle    = {International Conference on Machine Learning},
	publisher    = {PMLR},
	pages        = {10109--10126},
	issn         = {2640-3498},
	url          = {https://proceedings.mlr.press/v235/das24a.html},
	language     = {en}
}

@misc{deepseek-r1,
	title        = {DeepSeek-R1: Incentivizing Reasoning Capability in LLMs via Reinforcement Learning},
	author       = {DeepSeek-AI and Daya Guo and Dejian Yang and Haowei Zhang and Junxiao Song and Ruoyu Zhang and Runxin Xu and Qihao Zhu and Shirong Ma and Peiyi Wang and Xiao Bi and Xiaokang Zhang and Xingkai Yu and Yu Wu and Z. F. Wu and Zhibin Gou and Zhihong Shao and Zhuoshu Li and Ziyi Gao and others},
	year         = {2025},
	url          = {https://arxiv.org/abs/2501.12948},
	eprint       = {2501.12948},
	archiveprefix = {arXiv},
	primaryclass = {cs.CL}
}

@article{Devitt2017-ew,
	title        = {Episodic and semantic content of memory and imagination: A multilevel analysis},
	author       = {Devitt, Aleea L and Addis, Donna Rose and Schacter, Daniel L},
	year         = {2017},
	month        = oct,
	journal      = {Mem. Cognit.},
	publisher    = {Springer Science and Business Media LLC},
	volume       = {45},
	number       = {7},
	pages        = {1078--1094},
	language     = {en}
}

@article{KUMARAN2016512,
	title        = {What Learning Systems do Intelligent Agents Need? Complementary Learning Systems Theory Updated},
	author       = {Dharshan Kumaran and Demis Hassabis and James L. McClelland},
	year         = {2016},
	journal      = {Trends in Cognitive Sciences},
	volume       = {20},
	number       = {7},
	pages        = {512--534},
	doi          = {https://doi.org/10.1016/j.tics.2016.05.004},
	issn         = {1364-6613},
	url          = {https://www.sciencedirect.com/science/article/pii/S1364661316300432}
}

@article{motor,
	title        = {Reorganization and plasticity in the adult brain during learning of motor skills},
	author       = {Doyon, Julien and Benali, Habib},
	year         = {2005},
	month        = apr,
	journal      = {Curr. Opin. Neurobiol.},
	publisher    = {Elsevier BV},
	volume       = {15},
	number       = {2},
	pages        = {161--167},
	language     = {en}
}

@article{cortical-hippo,
	title        = {A cortical--hippocampal system for declarative memory},
	author       = {Eichenbaum, Howard},
	year         = {2000},
	journal      = {Nature Reviews Neuroscience},
	volume       = {1},
	number       = {1},
	pages        = {41--50},
	doi          = {10.1038/35036213},
	isbn         = {1471-0048},
	url          = {https://doi.org/10.1038/35036213},
	date         = {2000/10/01},
	id           = {Eichenbaum2000}
}

@article{Evans2024-pr,
	title        = {Declarative memory supports children's math skills: A longitudinal study},
	author       = {Evans, Tanya M and Lipscomb, Daniel W and Earle, F Sayako and Del Tufo, Stephanie N and Lum, Jarrad A G and Cutting, Laurie E and Ullman, Michael T},
	year         = {2024},
	month        = jul,
	journal      = {PLoS One},
	publisher    = {Public Library of Science (PLoS)},
	volume       = {19},
	number       = {7},
	pages        = {e0304211},
	copyright    = {http://creativecommons.org/licenses/by/4.0/},
	language     = {en}
}

@article{Fiser2002-aslin,
	title        = {Statistical learning of higher-order temporal structure from visual shape sequences},
	author       = {Fiser, J{\'o}zsef and Aslin, Richard N},
	year         = {2002},
	month        = may,
	journal      = {J. Exp. Psychol. Learn. Mem. Cogn.},
	publisher    = {American Psychological Association (APA)},
	volume       = {28},
	number       = {3},
	pages        = {458--467},
	language     = {en}
}

@article{Fiser2002-kt,
	title        = {Statistical learning of new visual feature combinations by infants},
	author       = {Fiser, J{\'o}zsef and Aslin, Richard N},
	year         = {2002},
	month        = nov,
	journal      = {Proc. Natl. Acad. Sci. U. S. A.},
	publisher    = {Proceedings of the National Academy of Sciences},
	volume       = {99},
	number       = {24},
	pages        = {15822--15826},
	language     = {en}
}

@article{Fleming2012-wm,
	title        = {The neural basis of metacognitive ability},
	author       = {Fleming, Stephen M and Dolan, Raymond J},
	year         = {2012},
	month        = may,
	journal      = {Philos. Trans. R. Soc. Lond. B Biol. Sci.},
	publisher    = {The Royal Society},
	volume       = {367},
	number       = {1594},
	pages        = {1338--1349},
	language     = {en}
}

@article{memory-activatioin-consolidation,
	title        = {The organization of recent and remote memories},
	author       = {Frankland, Paul W. and Bontempi, Bruno},
	year         = {2005},
	journal      = {Nature Reviews Neuroscience},
	volume       = {6},
	number       = {2},
	pages        = {119--130},
	doi          = {10.1038/nrn1607},
	isbn         = {1471-0048},
	url          = {https://doi.org/10.1038/nrn1607},
	date         = {2005/02/01},
	id           = {Frankland2005}
}

@article{Friedrich2013-ps,
	title        = {Mathematical logic in the human brain: semantics},
	author       = {Friedrich, Roland M and Friederici, Angela D},
	year         = {2013},
	month        = jan,
	journal      = {PLoS One},
	publisher    = {Public Library of Science (PLoS)},
	volume       = {8},
	number       = {1},
	pages        = {e53699},
	language     = {en}
}

@inproceedings{llm-not-abstract-reason,
	title        = {Large Language Models Are Not Strong Abstract Reasoners},
	author       = {Gendron, Gaël and Bao, Qiming and Witbrock, Michael and Dobbie, Gillian},
	year         = {2024},
	month        = {8},
	booktitle    = {Proceedings of the Thirty-Third International Joint Conference on Artificial Intelligence, {IJCAI-24}},
	publisher    = {International Joint Conferences on Artificial Intelligence Organization},
	pages        = {6270--6278},
	doi          = {10.24963/ijcai.2024/693},
	url          = {https://doi.org/10.24963/ijcai.2024/693},
	note         = {Main Track},
	editor       = {Kate Larson}
}

@article{implicit-dopamine,
	title        = {Modulation of striatal projection systems by dopamine},
	author       = {Gerfen, Charles R and Surmeier, D James},
	year         = {2011},
	month        = jul,
	journal      = {Annu. Rev. Neurosci.},
	publisher    = {Annual Reviews},
	volume       = {34},
	number       = {1},
	pages        = {441--466},
	language     = {en}
}

@article{Goode2020,
title = {An Integrated Index: Engrams, Place Cells, and Hippocampal Memory},
journal = {Neuron},
volume = {107},
number = {5},
pages = {805-820},
year = {2020},
issn = {0896-6273},
doi = {https://doi.org/10.1016/j.neuron.2020.07.011},
url = {https://www.sciencedirect.com/science/article/pii/S0896627320305286},
author = {Travis D. Goode and Kazumasa Z. Tanaka and Amar Sahay and Thomas J. McHugh}
}

@article{Goshen-cell,
	title        = {Dynamics of Retrieval Strategies for Remote Memories},
	author       = {Goshen, Inbal and Brodsky, Matthew and Prakash, Rohit and Wallace, Jenelle and Gradinaru, Viviana and Ramakrishnan, Charu and Deisseroth, Karl},
	year         = {2011},
	month        = oct,
	journal      = {Cell},
	publisher    = {Elsevier},
	volume       = {147},
	number       = {3},
	pages        = {678--689},
	doi          = {10.1016/j.cell.2011.09.033},
	isbn         = {0092-8674},
	url          = {https://doi.org/10.1016/j.cell.2011.09.033},
	date         = {2011/10/28},
	journal1     = {Cell},
	n2           = {Prevailing theory suggests that long-term memories are encoded via a two-phase process requiring early involvement of the hippocampus followed by the neocortex. Contextual fear memories in rodents rely on the hippocampus immediately following training but are unaffected by hippocampal lesions or pharmacological inhibition weeks later. With fast optogenetic methods, we examine the real-time contribution of hippocampal CA1 excitatory neurons to remote memory and find that contextual fear memory recall, even weeks after training, can be reversibly abolished by temporally precise optogenetic inhibition of CA1. When this inhibition is extended to match the typical time course of pharmacological inhibition, remote hippocampus dependence converts to hippocampus independence, suggesting that long-term memory retrieval normally depends on the hippocampus but can adaptively shift to alternate structures. Further revealing the plasticity of mechanisms required for memory recall, we confirm the remote-timescale importance of the anterior cingulate cortex (ACC) and implicate CA1 in ACC recruitment for remote recall.},
	type         = {doi: 10.1016/j.cell.2011.09.033},
	year1        = {2011}
}

@article{implicit-word-completion,
	title        = {Implicit and explicit memory for new associations in normal and amnesic subjects},
	author       = {Graf, Peter and Schacter, Daniel L},
	year         = {1985},
	journal      = {Journal of Experimental Psychology: Learning, Memory, and Cognition},
	publisher    = {American Psychological Association},
	address      = {US},
	volume       = {11},
	number       = {3},
	pages        = {501--518}
}

@article{habits-rituals,
	title        = {Habits, rituals, and the evaluative brain},
	author       = {Graybiel, Ann M},
	year         = {2008},
	month        = jul,
	journal      = {Annu. Rev. Neurosci.},
	publisher    = {Annual Reviews},
	volume       = {31},
	number       = {1},
	pages        = {359--387},
	language     = {en}
}

@article{Hampton2020-mf,
	title        = {Explicit memory and cognition in monkeys},
	author       = {Hampton, Robert R and Engelberg, Jonathan W M and Brady, Ryan J},
	year         = {2020},
	month        = feb,
	journal      = {Neuropsychologia},
	publisher    = {Elsevier BV},
	volume       = {138},
	number       = {107326},
	pages        = {107326},
	language     = {en}
}

@article{Hassabis2017-ul,
	title        = {{Neuroscience-Inspired} Artificial Intelligence},
	author       = {Hassabis, Demis and Kumaran, Dharshan and Summerfield, Christopher and Botvinick, Matthew},
	year         = {2017},
	month        = jul,
	journal      = {Neuron},
	address      = {United States},
	volume       = {95},
	number       = {2},
	pages        = {245--258},
	language     = {en}
}

@article{Hassabis2007-vo,
	title        = {Deconstructing episodic memory with construction},
	author       = {Hassabis, Demis and Maguire, Eleanor A},
	year         = {2007},
	month        = jul,
	journal      = {Trends Cogn. Sci.},
	publisher    = {Elsevier BV},
	volume       = {11},
	number       = {7},
	pages        = {299--306},
	language     = {en}
}

@article{basal-automatic,
	title        = {Switching from automatic to controlled behavior: cortico-basal ganglia mechanisms},
	author       = {Hikosaka, Okihide and Isoda, Masaki},
	year         = {2010},
	month        = apr,
	journal      = {Trends Cogn. Sci.},
	publisher    = {Elsevier BV},
	volume       = {14},
	number       = {4},
	pages        = {154--161},
	language     = {en}
}

@article{Irak2019-lc,
	title        = {Neurobiological basis of feeling of knowing in episodic memory},
	author       = {Irak, Metehan and Soylu, Can and Turan, G{\"o}zem and {\c C}apan, Dicle},
	year         = {2019},
	month        = jun,
	journal      = {Cogn. Neurodyn.},
	publisher    = {Springer Science and Business Media LLC},
	volume       = {13},
	number       = {3},
	pages        = {239--256},
	language     = {en}
}

@misc{emergent,
	title        = {Emergent Abilities of Large Language Models},
	author       = {Jason Wei and Yi Tay and Rishi Bommasani and Colin Raffel and Barret Zoph and Sebastian Borgeaud and Dani Yogatama and Maarten Bosma and Denny Zhou and Donald Metzler and Ed H. Chi and Tatsunori Hashimoto and Oriol Vinyals and Percy Liang and Jeff Dean and William Fedus},
	year         = {2022},
	url          = {https://arxiv.org/abs/2206.07682},
	eprint       = {2206.07682},
	archiveprefix = {arXiv},
	primaryclass = {cs.CL}
}

@inproceedings{huang2024large,
	title        = {Large Language Models Cannot Self-Correct Reasoning Yet},
	author       = {Jie Huang and Xinyun Chen and Swaroop Mishra and Huaixiu Steven Zheng and Adams Wei Yu and Xinying Song and Denny Zhou},
	year         = {2024},
	booktitle    = {The Twelfth International Conference on Learning Representations},
	url          = {https://openreview.net/forum?id=IkmD3fKBPQ}
}

@article{leutgeb2007pattern,
	title        = {Pattern Separation in the Dentate Gyrus and CA3 of the Hippocampus},
	author       = {Jill K. Leutgeb  and Stefan Leutgeb  and May-Britt Moser  and Edvard I. Moser},
	year         = {2007},
	journal      = {Science},
	volume       = {315},
	number       = {5814},
	pages        = {961--966},
	doi          = {10.1126/science.1135801},
	url          = {https://www.science.org/doi/abs/10.1126/science.1135801},
	eprint       = {https://www.science.org/doi/pdf/10.1126/science.1135801}
}

@article{Johns2012-om,
	title        = {Implicit and explicit olfactory memory in people with and without Down syndrome},
	author       = {Johns, Adam and Homewood, Judi and Stevenson, Richard and Taylor, Alan},
	year         = {2012},
	month        = mar,
	journal      = {Res. Dev. Disabil.},
	publisher    = {Elsevier BV},
	volume       = {33},
	number       = {2},
	pages        = {583--593},
	language     = {en}
}

@article{sparse-coding,
	title        = {Spatial selectivity of unit activity in the hippocampal granular layer},
	author       = {Jung, M W and McNaughton, B L},
	year         = {1993},
	month        = apr,
	journal      = {Hippocampus},
	publisher    = {Wiley},
	volume       = {3},
	number       = {2},
	pages        = {165--182},
	copyright    = {http://onlinelibrary.wiley.com/termsAndConditions\#vor},
	language     = {en}
}

@misc{llm-healthcare,
	title        = {A Survey of Large Language Models for Healthcare: from Data, Technology, and Applications to Accountability and Ethics},
	author       = {Kai He and Rui Mao and Qika Lin and Yucheng Ruan and Xiang Lan and Mengling Feng and Erik Cambria},
	year         = {2025},
	url          = {https://arxiv.org/abs/2310.05694},
	eprint       = {2310.05694},
	archiveprefix = {arXiv},
	primaryclass = {cs.CL}
}

@article{Klein2002-ce,
	title        = {Decisions and the evolution of memory: multiple systems, multiple functions},
	author       = {Klein, Stanley B and Cosmides, Leda and Tooby, John and Chance, Sarah},
	year         = {2002},
	month        = apr,
	journal      = {Psychol. Rev.},
	publisher    = {American Psychological Association (APA)},
	volume       = {109},
	number       = {2},
	pages        = {306--329},
	language     = {en}
}

@article{Klein2010-pb,
	title        = {The multiplicity of self: neuropsychological evidence and its implications for the self as a construct in psychological research},
	author       = {Klein, Stanley B and Gangi, Cynthia E},
	year         = {2010},
	month        = mar,
	journal      = {Ann. N. Y. Acad. Sci.},
	publisher    = {Wiley},
	volume       = {1191},
	number       = {1},
	pages        = {1--15},
	copyright    = {http://onlinelibrary.wiley.com/termsAndConditions\#vor},
	language     = {en}
}

@article{basal-gang-msn,
	title        = {Striatal Plasticity and Basal Ganglia Circuit Function},
	author       = {Kreitzer, Anatol C. and Malenka, Robert C.},
	year         = {2008},
	month        = nov,
	journal      = {Neuron},
	publisher    = {Elsevier},
	volume       = {60},
	number       = {4},
	pages        = {543--554},
	doi          = {10.1016/j.neuron.2008.11.005},
	isbn         = {0896-6273},
	url          = {https://doi.org/10.1016/j.neuron.2008.11.005},
	date         = {2008/11/26},
	journal1     = {Neuron},
	type         = {doi: 10.1016/j.neuron.2008.11.005},
	year1        = {2008}
}

@article{bio-underping,
	title        = {Biological underpinnings for lifelong learning machines},
	author       = {Kudithipudi, Dhireesha and Aguilar-Simon, Mario and Babb, Jonathan and Bazhenov, Maxim and Blackiston, Douglas and Bongard, Josh and Brna, Andrew P and Chakravarthi Raja, Suraj and Cheney, Nick and Clune, Jeff and Daram, Anurag and Fusi, Stefano and Helfer, Peter and Kay, Leslie and Ketz, Nicholas and Kira, Zsolt and Kolouri, Soheil and Krichmar, Jeffrey L and Kriegman, Sam and Levin, Michael and Madireddy, Sandeep and Manicka, Santosh and Marjaninejad, Ali and McNaughton, Bruce and Miikkulainen, Risto and Navratilova, Zaneta and Pandit, Tej and Parker, Alice and Pilly, Praveen K and Risi, Sebastian and Sejnowski, Terrence J and Soltoggio, Andrea and Soures, Nicholas and Tolias, Andreas S and Urbina-Meléndez, Darío and Valero-Cuevas, Francisco J and van de Ven, Gido M and Vogelstein, Joshua T and Wang, Felix and Weiss, Ron and Yanguas-Gil, Angel and Zou, Xinyun and Siegelmann, Hava},
	year         = {2022},
	month        = mar,
	journal      = {Nature machine intelligence},
	publisher    = {Springer Science and Business Media LLC},
	volume       = {4},
	number       = {3},
	pages        = {196--210},
	doi          = {10.1038/s42256-022-00452-0},
	issn         = {2522-5839,2522-5839},
	url          = {https://www.nature.com/articles/s42256-022-00452-0},
	language     = {en}
}

@article{SQUIRE2004171,
	title        = {Memory systems of the brain: A brief history and current perspective},
	author       = {Larry R. Squire},
	year         = {2004},
	journal      = {Neurobiology of Learning and Memory},
	volume       = {82},
	number       = {3},
	pages        = {171--177},
	doi          = {https://doi.org/10.1016/j.nlm.2004.06.005},
	issn         = {1074-7427},
	url          = {https://www.sciencedirect.com/science/article/pii/S1074742704000735},
	note         = {Multiple Memory Systems}
}

@article{declarative-and-nondeclarative,
	title        = {Structure and function of declarative and nondeclarative memory systems},
	author       = {Larry R. Squire  and Stuart M. Zola},
	year         = {1996},
	journal      = {Proceedings of the National Academy of Sciences},
	volume       = {93},
	number       = {24},
	pages        = {13515--13522},
	doi          = {10.1073/pnas.93.24.13515},
	url          = {https://www.pnas.org/doi/abs/10.1073/pnas.93.24.13515},
	eprint       = {https://www.pnas.org/doi/pdf/10.1073/pnas.93.24.13515}
}

@article{BATTERINK201731,
	title        = {Online neural monitoring of statistical learning},
	author       = {Laura J. Batterink and Ken A. Paller},
	year         = {2017},
	journal      = {Cortex},
	volume       = {90},
	pages        = {31--45},
	doi          = {https://doi.org/10.1016/j.cortex.2017.02.004},
	issn         = {0010-9452},
	url          = {https://www.sciencedirect.com/science/article/pii/S0010945217300540}
}

@article{Lieberman1992-lv,
	title        = {Speech production, syntax comprehension, and cognitive deficits in Parkinson's disease},
	author       = {Lieberman, Philip and Kako, Edward and Friedman, Joseph and Tajchman, Gary and Feldman, Liane S and Jiminez, Elsa B},
	year         = {1992},
	month        = aug,
	journal      = {Brain Lang.},
	publisher    = {Elsevier BV},
	volume       = {43},
	number       = {2},
	pages        = {169--189},
	copyright    = {http://creativecommons.org/licenses/by-nc-nd/4.0/},
	language     = {en}
}

@article{schrimpf2021neural,
	title        = {The neural architecture of language: Integrative modeling converges on predictive processing},
	author       = {Martin Schrimpf  and Idan Asher Blank  and Greta Tuckute  and Carina Kauf  and Eghbal A. Hosseini  and Nancy Kanwisher  and Joshua B. Tenenbaum  and Evelina Fedorenko},
	year         = {2021},
	journal      = {Proceedings of the National Academy of Sciences},
	volume       = {118},
	number       = {45},
	pages        = {e2105646118},
	doi          = {10.1073/pnas.2105646118},
	url          = {https://www.pnas.org/doi/abs/10.1073/pnas.2105646118},
	eprint       = {https://www.pnas.org/doi/pdf/10.1073/pnas.2105646118}
}

@article{Mattar2018,
	title        = {Prioritized memory access explains planning and hippocampal replay},
	author       = {Mattar, Marcelo G. and Daw, Nathaniel D.},
	year         = {2018},
	journal      = {Nature Neuroscience},
	volume       = {21},
	number       = {11},
	pages        = {1609--1617},
	doi          = {10.1038/s41593-018-0232-z},
	isbn         = {1546-1726},
	url          = {https://doi.org/10.1038/s41593-018-0232-z},
	date         = {2018/11/01},
	id           = {Mattar2018}
}

@article{hippo-neo-cotex,
	title        = {Why there are complementary learning systems in the hippocampus and neocortex: insights from the successes and failures of connectionist models of learning and memory},
	author       = {McClelland, James L and McNaughton, Bruce L and O'Reilly, Randall C},
	year         = {1995},
	month        = jul,
	journal      = {Psychol. Rev.},
	publisher    = {American Psychological Association (APA)},
	volume       = {102},
	number       = {3},
	pages        = {419--457},
	language     = {en}
}

@incollection{Menon2016-nr,
	title        = {Memory and cognitive control circuits in mathematical cognition and learning},
	author       = {Menon, V},
	year         = {2016},
	booktitle    = {Progress in Brain Research},
	publisher    = {Elsevier},
	series       = {Progress in brain research},
	pages        = {159--186}
}

@article{FAYOL2012392,
	title        = {The use of procedural knowledge in simple addition and subtraction problems},
	author       = {Michel Fayol and Catherine Thevenot},
	year         = {2012},
	journal      = {Cognition},
	volume       = {123},
	number       = {3},
	pages        = {392--403},
	doi          = {https://doi.org/10.1016/j.cognition.2012.02.008},
	issn         = {0010-0277},
	url          = {https://www.sciencedirect.com/science/article/pii/S0010027712000340}
}

@inproceedings{turpin2023language,
	title        = {Language Models Don't Always Say What They Think: Unfaithful Explanations in Chain-of-Thought Prompting},
	author       = {Miles Turpin and Julian Michael and Ethan Perez and Samuel R. Bowman},
	year         = {2023},
	booktitle    = {Thirty-seventh Conference on Neural Information Processing Systems},
	url          = {https://openreview.net/forum?id=bzs4uPLXvi}
}

@article{dopamine-computation,
	title        = {Computational roles for dopamine in behavioural control},
	author       = {Montague, P. Read and Hyman, Steven E. and Cohen, Jonathan D.},
	year         = {2004},
	journal      = {Nature},
	volume       = {431},
	number       = {7010},
	pages        = {760--767},
	doi          = {10.1038/nature03015},
	isbn         = {1476-4687},
	url          = {https://doi.org/10.1038/nature03015},
	date         = {2004/10/01},
	id           = {Montague2004}
}

@inproceedings{levels-of-agi,
	title        = {Position: Levels of {AGI} for Operationalizing Progress on the Path to {AGI}},
	author       = {Morris, Meredith Ringel and Sohl-Dickstein, Jascha and Fiedel, Noah and Warkentin, Tris and Dafoe, Allan and Faust, Aleksandra and Farabet, Clement and Legg, Shane},
	year         = {2024},
	month        = {21--27 Jul},
	booktitle    = {Proceedings of the 41st International Conference on Machine Learning},
	publisher    = {PMLR},
	series       = {Proceedings of Machine Learning Research},
	volume       = {235},
	pages        = {36308--36321},
	url          = {https://proceedings.mlr.press/v235/morris24b.html},
	editor       = {Salakhutdinov, Ruslan and Kolter, Zico and Heller, Katherine and Weller, Adrian and Oliver, Nuria and Scarlett, Jonathan and Berkenkamp, Felix}
}

@article{ca1-hippo-output,
	title        = {NMDA receptors, place cells and hippocampal spatial memory},
	author       = {Nakazawa, Kazu and McHugh, Thomas J. and Wilson, Matthew A. and Tonegawa, Susumu},
	year         = {2004},
	journal      = {Nature Reviews Neuroscience},
	volume       = {5},
	number       = {5},
	pages        = {361--372},
	doi          = {10.1038/nrn1385},
	isbn         = {1471-0048},
	url          = {https://doi.org/10.1038/nrn1385},
	date         = {2004/05/01},
	id           = {Nakazawa2004}
}

@article{ca3-nmda,
	title        = {Requirement for hippocampal {CA3} {NMDA} receptors in associative memory recall},
	author       = {Nakazawa, Kazu and Quirk, Michael C and Chitwood, Raymond A and Watanabe, Masahiko and Yeckel, Mark F and Sun, Linus D and Kato, Akira and Carr, Candice A and Johnston, Daniel and Wilson, Matthew A and Tonegawa, Susumu},
	year         = {2002},
	month        = jul,
	journal      = {Science},
	publisher    = {American Association for the Advancement of Science (AAAS)},
	volume       = {297},
	number       = {5579},
	pages        = {211--218},
	language     = {en}
}

@article{implicit-dissociation,
	title        = {Preserved Learning and Retention of Pattern-Analyzing Skill in Amnesia: Dissociation of Knowing How and Knowing That},
	author       = {Neal J. Cohen  and Larry R. Squire},
	year         = {1980},
	journal      = {Science},
	volume       = {210},
	number       = {4466},
	pages        = {207--210},
	doi          = {10.1126/science.7414331},
	url          = {https://www.science.org/doi/abs/10.1126/science.7414331},
	eprint       = {https://www.science.org/doi/pdf/10.1126/science.7414331}
}

@article{implicit-basal,
	title        = {Learning and memory functions of the Basal Ganglia},
	author       = {Packard, Mark G and Knowlton, Barbara J},
	year         = {2002},
	month        = mar,
	journal      = {Annu. Rev. Neurosci.},
	publisher    = {Annual Reviews},
	volume       = {25},
	number       = {1},
	pages        = {563--593},
	language     = {en}
}

@inproceedings{memoria,
	title        = {Memoria: Resolving Fateful Forgetting Problem through Human-Inspired Memory Architecture},
	author       = {Park, Sangjun and Bak, Jinyeong},
	year         = {2024},
	month        = {21--27 Jul},
	booktitle    = {Proceedings of the 41st International Conference on Machine Learning},
	publisher    = {PMLR},
	series       = {Proceedings of Machine Learning Research},
	volume       = {235},
	pages        = {39587--39615},
	url          = {https://proceedings.mlr.press/v235/park24a.html},
	editor       = {Salakhutdinov, Ruslan and Kolter, Zico and Heller, Katherine and Weller, Adrian and Oliver, Nuria and Scarlett, Jonathan and Berkenkamp, Felix}
}

@article{semantic-mem,
	title        = {Where do you know what you know? The representation of semantic knowledge in the human brain},
	author       = {Patterson, Karalyn and Nestor, Peter J. and Rogers, Timothy T.},
	year         = {2007},
	journal      = {Nature Reviews Neuroscience},
	volume       = {8},
	number       = {12},
	pages        = {976--987},
	doi          = {10.1038/nrn2277},
	isbn         = {1471-0048},
	url          = {https://doi.org/10.1038/nrn2277},
	date         = {2007/12/01},
	id           = {Patterson2007}
}

@article{Pedraza-24,
	title        = {Evidence for a competitive relationship between executive functions and statistical learning},
	author       = {Pedraza, Felipe and Farkas, Bence C. and V{\'e}kony, Teod{\'o}ra and Haesebaert, Frederic and Phelipon, Romane and Mihalecz, Imola and Janacsek, Karolina and Anders, Royce and Tillmann, Barbara and Plancher, Ga{\"e}n and N{\'e}meth, Dezs{\H o}},
	year         = {2024},
	journal      = {npj Science of Learning},
	volume       = {9},
	number       = {1},
	pages        = {30},
	doi          = {10.1038/s41539-024-00243-9},
	isbn         = {2056-7936},
	url          = {https://doi.org/10.1038/s41539-024-00243-9},
	date         = {2024/04/12},
	id           = {Pedraza2024}
}

@misc{helm,
	title        = {Holistic Evaluation of Language Models},
	author       = {Percy Liang and Rishi Bommasani and Tony Lee and Dimitris Tsipras and Dilara Soylu and Michihiro Yasunaga and Yian Zhang and Deepak Narayanan and Yuhuai Wu and Ananya Kumar and Benjamin Newman and Binhang Yuan and Bobby Yan and Ce Zhang and Christian Cosgrove and Christopher D. Manning and Christopher Ré and Diana Acosta-Navas and Drew A. Hudson and others},
	year         = {2023},
	url          = {https://arxiv.org/abs/2211.09110},
	eprint       = {2211.09110},
	archiveprefix = {arXiv},
	primaryclass = {cs.CL}
}

@misc{llm-law,
	title        = {SaulLM-7B: A pioneering Large Language Model for Law},
	author       = {Pierre Colombo and Telmo Pessoa Pires and Malik Boudiaf and Dominic Culver and Rui Melo and Caio Corro and Andre F. T. Martins and Fabrizio Esposito and Vera Lúcia Raposo and Sofia Morgado and Michael Desa},
	year         = {2024},
	url          = {https://arxiv.org/abs/2403.03883},
	eprint       = {2403.03883},
	archiveprefix = {arXiv},
	primaryclass = {cs.CL}
}

@incollection{human-memory-atkinson,
	title        = {Human Memory: A Proposed System and its Control Processes},
	author       = {R.C. Atkinson and R.M. Shiffrin},
	year         = {1968},
	publisher    = {Academic Press},
	series       = {Psychology of Learning and Motivation},
	volume       = {2},
	pages        = {89--195},
	doi          = {https://doi.org/10.1016/S0079-7421(08)60422-3},
	issn         = {0079-7421},
	url          = {https://www.sciencedirect.com/science/article/pii/S0079742108604223},
	editor       = {Kenneth W. Spence and Janet Taylor Spence}
}

@article{comparison-memories,
	title        = {The neural basis of implicit learning and memory: a review of neuropsychological and neuroimaging research},
	author       = {Reber, Paul J},
	year         = {2013},
	month        = aug,
	journal      = {Neuropsychologia},
	publisher    = {Elsevier BV},
	volume       = {51},
	number       = {10},
	pages        = {2026--2042},
	language     = {en}
}

@article{ca3-circuit,
	title        = {Operation and plasticity of hippocampal {CA3} circuits: implications for memory encoding},
	author       = {Rebola, Nelson and Carta, Mario and Mulle, Christophe},
	year         = {2017},
	month        = apr,
	journal      = {Nat. Rev. Neurosci.},
	publisher    = {Springer Science and Business Media LLC},
	volume       = {18},
	number       = {4},
	pages        = {208--220},
	language     = {en}
}

@article{ca1-ltp-ltd,
	title        = {LTP and LTD: An Embarrassment of Riches},
	author       = {Robert C. Malenka and Mark F. Bear},
	year         = {2004},
	journal      = {Neuron},
	volume       = {44},
	number       = {1},
	pages        = {5--21},
	doi          = {https://doi.org/10.1016/j.neuron.2004.09.012},
	issn         = {0896-6273},
	url          = {https://www.sciencedirect.com/science/article/pii/S0896627304006087}
}

@article{pattern-comp-sep,
	title        = {The mechanisms for pattern completion and pattern separation in the hippocampus},
	author       = {Rolls, Edmund},
	year         = {2013},
	journal      = {Frontiers in Systems Neuroscience},
	volume       = {7},
	doi          = {10.3389/fnsys.2013.00074},
	issn         = {1662-5137},
	url          = {https://www.frontiersin.org/journals/systems-neuroscience/articles/10.3389/fnsys.2013.00074}
}

@misc{kadavath2022languagemodelsmostlyknow,
	title        = {Language Models (Mostly) Know What They Know},
	author       = {Saurav Kadavath and Tom Conerly and Amanda Askell and Tom Henighan and Dawn Drain and Ethan Perez and Nicholas Schiefer and Zac Hatfield-Dodds and Nova DasSarma and Eli Tran-Johnson and Scott Johnston and Sheer El-Showk and Andy Jones and Nelson Elhage and Tristan Hume and Anna Chen and Yuntao Bai and Sam Bowman and Stanislav Fort and Deep Ganguli and Danny Hernandez and Josh Jacobson and Jackson Kernion and Shauna Kravec and Liane Lovitt and Kamal Ndousse and Catherine Olsson and Sam Ringer and Dario Amodei and Tom Brown and Jack Clark and Nicholas Joseph and Ben Mann and Sam McCandlish and Chris Olah and Jared Kaplan},
	year         = {2022},
	url          = {https://arxiv.org/abs/2207.05221},
	eprint       = {2207.05221},
	archiveprefix = {arXiv},
	primaryclass = {cs.CL}
}

@article{implicit-mem,
	title        = {Implicit memory: History and current status},
	author       = {Schacter, Daniel L},
	year         = {1987},
	journal      = {Journal of Experimental Psychology: Learning, Memory, and Cognition},
	publisher    = {American Psychological Association},
	address      = {US},
	volume       = {13},
	number       = {3},
	pages        = {501--518}
}

@article{episodic-non-illusion,
	title        = {The cognitive neuroscience of constructive memory: remembering the past and imagining the future},
	author       = {Schacter, Daniel L and Addis, Donna Rose},
	year         = {2007},
	month        = may,
	journal      = {Philos. Trans. R. Soc. Lond. B Biol. Sci.},
	publisher    = {The Royal Society},
	volume       = {362},
	number       = {1481},
	pages        = {773--786},
	language     = {en}
}

@article{Schacter2017-vy,
	title        = {Episodic future thinking: mechanisms and functions},
	author       = {Schacter, Daniel L and Benoit, Roland G and Szpunar, Karl K},
	year         = {2017},
	month        = oct,
	journal      = {Curr. Opin. Behav. Sci.},
	publisher    = {Elsevier BV},
	volume       = {17},
	pages        = {41--50}
}

@article{Schlegel-nature,
	title        = {Whole-brain annotation and multi-connectome cell typing of Drosophila},
	author       = {Schlegel, Philipp and Yin, Yijie and Bates, Alexander S. and Dorkenwald, Sven and Eichler, Katharina and Brooks, Paul and Han, Daniel S. and Gkantia, Marina and dos Santos, Marcia and Munnelly, Eva J. and Badalamente, Griffin and Serratosa Capdevila, Laia and Sane, Varun A. and Fragniere, Alexandra M. C. and Kiassat, Ladann and Pleijzier, Markus W. and St{\"u}rner, Tomke and Tamimi, Imaan F. M. and Dunne, Christopher R. and others},
	year         = {2024},
	journal      = {Nature},
	volume       = {634},
	number       = {8032},
	pages        = {139--152},
	doi          = {10.1038/s41586-024-07686-5},
	isbn         = {1476-4687},
	url          = {https://doi.org/10.1038/s41586-024-07686-5},
	date         = {2024/10/01},
	id           = {Schlegel2024}
}

@article{Schmolck2002-ul--hm,
	title        = {Semantic knowledge in patient {H.M}. and other patients with bilateral medial and lateral temporal lobe lesions},
	author       = {Schmolck, Heike and Kensinger, Elizabeth A and Corkin, Suzanne and Squire, Larry R},
	year         = {2002},
	journal      = {Hippocampus},
	publisher    = {Wiley},
	volume       = {12},
	number       = {4},
	pages        = {520--533},
	copyright    = {http://onlinelibrary.wiley.com/termsAndConditions\#vor},
	language     = {en}
}

@article{hm-hippocampus,
	title        = {Loss of recent memory after bilateral hippocampal lesions},
	author       = {Scoville, W B and Milner, B},
	year         = {1957},
	month        = feb,
	journal      = {J. Neurol. Neurosurg. Psychiatry},
	publisher    = {BMJ},
	volume       = {20},
	number       = {1},
	pages        = {11--21},
	language     = {en}
}

@article{ai-native,
	title        = {{AI}-native memory: A pathway from {LLMs} towards {AGI}},
	author       = {Shang, Jingbo and Zheng, Zai and Wei, Jiale and Ying, Xiang and Tao, Felix and {Mindverse Team}},
	year         = {2024},
	month        = jun,
	journal      = {arXiv [cs.CL]},
	url          = {http://arxiv.org/abs/2406.18312},
	archiveprefix = {arXiv},
	primaryclass = {cs.CL},
	eprint       = {2406.18312}
}

@inproceedings{llm-cog-failures,
	title        = {A Survey on Large Language Model Reasoning Failures},
	author       = {Song, Peiyang and Han, Pengrui and Goodman, Noah},
	year         = {2025},
	month        = jul,
	booktitle    = {2nd AI for Math Workshop @ ICML 2025},
	url          = {https://openreview.net/forum?id=hsgMn4KBFG}
}

@article{medial-temporal-memory,
	title        = {The medial temporal lobe memory system},
	author       = {Squire, L R and Zola-Morgan, S},
	year         = {1991},
	month        = sep,
	journal      = {Science},
	publisher    = {American Association for the Advancement of Science (AAAS)},
	volume       = {253},
	number       = {5026},
	pages        = {1380--1386},
	language     = {en}
}

@article{implicit-circuit,
	title        = {Conscious and unconscious memory systems},
	author       = {Squire, Larry R and Dede, Adam J O},
	year         = {2015},
	month        = mar,
	journal      = {Cold Spring Harb. Perspect. Biol.},
	publisher    = {Cold Spring Harbor Laboratory},
	volume       = {7},
	number       = {3},
	pages        = {a021667},
	language     = {en}
}

@article{doi:10.1126/science.1239073,
	title        = {Creating a False Memory in the Hippocampus},
	author       = {Steve Ramirez  and Xu Liu  and Pei-Ann Lin  and Junghyup Suh  and Michele Pignatelli  and Roger L. Redondo  and Tomás J. Ryan  and Susumu Tonegawa},
	year         = {2013},
	journal      = {Science},
	volume       = {341},
	number       = {6144},
	pages        = {387--391},
	doi          = {10.1126/science.1239073},
	url          = {https://www.science.org/doi/abs/10.1126/science.1239073},
	eprint       = {https://www.science.org/doi/pdf/10.1126/science.1239073}
}

@article{Suddendorf2007-dh,
	title        = {The evolution of foresight: What is mental time travel, and is it unique to humans?},
	author       = {Suddendorf, Thomas and Corballis, Michael C},
	year         = {2007},
	month        = jun,
	journal      = {Behav. Brain Sci.},
	publisher    = {Cambridge University Press (CUP)},
	volume       = {30},
	number       = {3},
	pages        = {299--313; discussion 313--51},
	language     = {en}
}

@article{Szigeti2014-de,
	title        = {{OpenWorm}: an open-science approach to modeling Caenorhabditis elegans},
	author       = {Szigeti, Bal{\'a}zs and Gleeson, Padraig and Vella, Michael and Khayrulin, Sergey and Palyanov, Andrey and Hokanson, Jim and Currie, Michael and Cantarelli, Matteo and Idili, Giovanni and Larson, Stephen},
	year         = {2014},
	month        = nov,
	journal      = {Front Comput Neurosci},
	address      = {Switzerland},
	volume       = {8},
	pages        = {137},
	language     = {en}
}

@article{how-far-agi-llm,
	title        = {How Far Are We From {AGI}: Are {LLM}s All We Need?},
	author       = {Tao Feng and Chuanyang Jin and Jingyu Liu and Kunlun Zhu and Haoqin Tu and Zirui Cheng and Guanyu Lin and Jiaxuan You},
	year         = {2024},
	journal      = {Transactions on Machine Learning Research},
	issn         = {2835-8856},
	url          = {https://openreview.net/forum?id=H2ZKqfNd0U},
	note         = {Survey Certification}
}

@article{hipo.20350,
	title        = {The hippocampal indexing theory and episodic memory: Updating the index},
	author       = {Teyler, Timothy J. and Rudy, Jerry W.},
	year         = {2007},
	journal      = {Hippocampus},
	volume       = {17},
	number       = {12},
	pages        = {1158--1169},
	doi          = {https://doi.org/10.1002/hipo.20350},
	url          = {https://onlinelibrary.wiley.com/doi/abs/10.1002/hipo.20350},
	eprint       = {https://onlinelibrary.wiley.com/doi/pdf/10.1002/hipo.20350}
}

@incollection{NELSON1990125,
	title        = {Metamemory: A Theoretical Framework and New Findings},
	author       = {Thomas O. Nelson},
	year         = {1990},
	publisher    = {Academic Press},
	series       = {Psychology of Learning and Motivation},
	volume       = {26},
	pages        = {125--173},
	doi          = {https://doi.org/10.1016/S0079-7421(08)60053-5},
	issn         = {0079-7421},
	url          = {https://www.sciencedirect.com/science/article/pii/S0079742108600535},
	editor       = {Gordon H. Bower}
}

@inproceedings{llm-creativity,
	title        = {{M}ac{G}yver: Are Large Language Models Creative Problem Solvers?},
	author       = {Tian, Yufei  and Ravichander, Abhilasha  and Qin, Lianhui  and Le Bras, Ronan  and Marjieh, Raja  and Peng, Nanyun  and Choi, Yejin  and Griffiths, Thomas  and Brahman, Faeze},
	year         = {2024},
	month        = jun,
	booktitle    = {Proceedings of the 2024 Conference of the North American Chapter of the Association for Computational Linguistics: Human Language Technologies (Volume 1: Long Papers)},
	publisher    = {Association for Computational Linguistics},
	address      = {Mexico City, Mexico},
	pages        = {5303--5324},
	doi          = {10.18653/v1/2024.naacl-long.297},
	url          = {https://aclanthology.org/2024.naacl-long.297/},
	editor       = {Duh, Kevin  and Gomez, Helena  and Bethard, Steven}
}

@article{dopamine-modulation,
	title        = {Dopaminergic modulation of synaptic transmission in cortex and striatum},
	author       = {Tritsch, Nicolas X and Sabatini, Bernardo L},
	year         = {2012},
	month        = oct,
	journal      = {Neuron},
	publisher    = {Elsevier BV},
	volume       = {76},
	number       = {1},
	pages        = {33--50},
	copyright    = {https://www.elsevier.com/open-access/userlicense/1.0/},
	language     = {en}
}

@article{Tulving-episodic-and-semantic,
	title        = {Episodic and semantic memory},
	author       = {Tulving, Endel},
	year         = {1972},
	journal      = {Organization of memory.},
	publisher    = {Academic Press},
	address      = {Oxford, England},
	pages        = {xiii, 423--xiii, 423}
}

@article{tulving-memory-consc,
	title        = {Memory and consciousness},
	author       = {Tulving, Endel},
	year         = {1985},
	journal      = {Canadian Psychology / Psychologie canadienne},
	publisher    = {Canadian Psychological Association},
	address      = {Canada},
	volume       = {26},
	number       = {1},
	pages        = {1--12}
}

@article{tulving-epi-mem,
	title        = {Episodic Memory: From Mind to Brain},
	author       = {Tulving, Endel},
	year         = {2002},
	journal      = {Annual Review of Psychology},
	publisher    = {Annual Reviews},
	volume       = {53},
	number       = {Volume 53, 2002},
	pages        = {1--25},
	doi          = {https://doi.org/10.1146/annurev.psych.53.100901.135114},
	issn         = {1545-2085},
	url          = {https://www.annualreviews.org/content/journals/10.1146/annurev.psych.53.100901.135114},
	type         = {Journal Article}
}

@article{Turk-Browne2005-tu,
	title        = {The automaticity of visual statistical learning},
	author       = {Turk-Browne, Nicholas B and Jung{\'e}, Justin and Scholl, Brian J},
	year         = {2005},
	month        = nov,
	journal      = {J. Exp. Psychol. Gen.},
	publisher    = {American Psychological Association (APA)},
	volume       = {134},
	number       = {4},
	pages        = {552--564},
	language     = {en}
}

@article{Ullman1997-dy,
	title        = {A neural dissociation within language: Evidence that the mental dictionary is part of declarative memory, and that grammatical rules are processed by the procedural system},
	author       = {Ullman, M T and Corkin, S and Coppola, M and Hickok, G and Growdon, J H and Koroshetz, W J and Pinker, S},
	year         = {1997},
	month        = mar,
	journal      = {J. Cogn. Neurosci.},
	publisher    = {MIT Press - Journals},
	volume       = {9},
	number       = {2},
	pages        = {266--276},
	language     = {en}
}

@article{Ullman2004-hj,
	title        = {Contributions of memory circuits to language: the declarative/procedural model},
	author       = {Ullman, Michael T},
	year         = {2004},
	month        = may,
	journal      = {Cognition},
	publisher    = {Elsevier BV},
	volume       = {92},
	number       = {1-2},
	pages        = {231--270},
	language     = {en}
}

@inproceedings{llm-still-cant-plan,
	title        = {Large Language Models Still Can't Plan (A Benchmark for {LLMs} on Planning and Reasoning about Change)},
	author       = {Valmeekam, Karthik and Olmo, Alberto and Sreedharan, Sarath and Kambhampati, Subbarao},
	year         = {2022},
	month        = nov,
	booktitle    = {NeurIPS 2022 Foundation Models for Decision Making Workshop},
	url          = {https://openreview.net/pdf?id=wUU-7XTL5XO}
}

@article{Vargha-Khadem1997-jj,
	title        = {Differential effects of early hippocampal pathology on episodic and semantic memory},
	author       = {Vargha-Khadem, F and Gadian, D G and Watkins, K E and Connelly, A and Van Paesschen, W and Mishkin, M},
	year         = {1997},
	month        = jul,
	journal      = {Science},
	publisher    = {American Association for the Advancement of Science (AAAS)},
	volume       = {277},
	number       = {5324},
	pages        = {376--380},
	language     = {en}
}

@inproceedings{NEURIPS2023_ebd82705,
	title        = {Augmenting Language Models with Long-Term Memory},
	author       = {Wang, Weizhi and Dong, Li and Cheng, Hao and Liu, Xiaodong and Yan, Xifeng and Gao, Jianfeng and Wei, Furu},
	year         = {2023},
	booktitle    = {Advances in Neural Information Processing Systems},
	publisher    = {Curran Associates, Inc.},
	volume       = {36},
	pages        = {74530--74543},
	url          = {https://proceedings.neurips.cc/paper_files/paper/2023/file/ebd82705f44793b6f9ade5a669d0f0bf-Paper-Conference.pdf},
	editor       = {A. Oh and T. Naumann and A. Globerson and K. Saenko and M. Hardt and S. Levine}
}

@inproceedings{memoryllm,
	title        = {{MEMORYLLM}: Towards Self-Updatable Large Language Models},
	author       = {Wang, Yu and Gao, Yifan and Chen, Xiusi and Jiang, Haoming and Li, Shiyang and Yang, Jingfeng and Yin, Qingyu and Li, Zheng and Li, Xian and Yin, Bing and Shang, Jingbo and Mcauley, Julian},
	year         = {2024},
	month        = {21--27 Jul},
	booktitle    = {Proceedings of the 41st International Conference on Machine Learning},
	publisher    = {PMLR},
	series       = {Proceedings of Machine Learning Research},
	volume       = {235},
	pages        = {50453--50466},
	url          = {https://proceedings.mlr.press/v235/wang24s.html},
	editor       = {Salakhutdinov, Ruslan and Kolter, Zico and Heller, Katherine and Weller, Adrian and Oliver, Nuria and Scarlett, Jonathan and Berkenkamp, Felix}
}

@article{Whittington-cell,
	title        = {The Tolman-Eichenbaum Machine: Unifying Space and Relational Memory through Generalization in the Hippocampal Formation},
	author       = {Whittington, James C. R. and Muller, Timothy H. and Mark, Shirley and Chen, Guifen and Barry, Caswell and Burgess, Neil and Behrens, Timothy E. J.},
	year         = {2020},
	month        = nov,
	journal      = {Cell},
	publisher    = {Elsevier},
	volume       = {183},
	number       = {5},
	pages        = {1249--1263.e23},
	doi          = {10.1016/j.cell.2020.10.024},
	isbn         = {0092-8674},
	url          = {https://doi.org/10.1016/j.cell.2020.10.024},
	date         = {2020/11/25},
	journal1     = {Cell},
	n2           = {The hippocampal-entorhinal system is important for spatial and relational memory tasks. We formally link these domains, provide a mechanistic understanding of the hippocampal role in generalization, and offer unifying principles underlying many entorhinal and hippocampal cell types. We propose medial entorhinal cells form a basis describing structural knowledge, and hippocampal cells link this basis with sensory representations. Adopting these principles, we introduce the Tolman-Eichenbaum machine (TEM). After learning, TEM entorhinal cells display diverse properties resembling apparently bespoke spatial responses, such as grid, band, border, and object-vector cells. TEM hippocampal cells include place and landmark cells that remap between environments. Crucially, TEM also aligns with empirically recorded representations in complex non-spatial tasks. TEM also generates predictions that hippocampal remapping is not random as previously believed; rather, structural knowledge is preserved across environments. We confirm this structural transfer over remapping in simultaneously recorded place and grid cells.},
	type         = {doi: 10.1016/j.cell.2020.10.024},
	year1        = {2020}
}

@article{hippo-retrieval,
	title        = {The hippocampus plays a selective role in the retrieval of detailed contextual memories},
	author       = {Wiltgen, Brian J and Zhou, Miou and Cai, Ying and Balaji, J and Karlsson, Mikael Guzman and Parivash, Sherveen N and Li, Weidong and Silva, Alcino J},
	year         = {2010},
	month        = aug,
	journal      = {Curr. Biol.},
	publisher    = {Elsevier BV},
	volume       = {20},
	number       = {15},
	pages        = {1336--1344},
	copyright    = {https://www.elsevier.com/open-access/userlicense/1.0/},
	language     = {en}
}

@article{prediction-and-reward,
	title        = {A Neural Substrate of Prediction and Reward},
	author       = {Wolfram Schultz  and Peter Dayan  and P. Read Montague},
	year         = {1997},
	journal      = {Science},
	volume       = {275},
	number       = {5306},
	pages        = {1593--1599},
	doi          = {10.1126/science.275.5306.1593},
	url          = {https://www.science.org/doi/abs/10.1126/science.275.5306.1593},
	eprint       = {https://www.science.org/doi/pdf/10.1126/science.275.5306.1593}
}

@inproceedings{Xiao2024-au,
	title        = {{InfLLM}: Training-Free Long-Context Extrapolation for {LLMs} with an Efficient Context Memory},
	author       = {Xiao, Chaojun and Zhang, Pengle and Han, Xu and Xiao, Guangxuan and Lin, Yankai and Zhang, Zhengyan and Liu, Zhiyuan and Sun, Maosong},
	year         = {2024},
	month        = nov,
	booktitle    = {The Thirty-eighth Annual Conference on Neural Information Processing Systems},
	url          = {https://openreview.net/pdf?id=bTHFrqhASY}
}

@article{Yang2024-io,
	title        = {$\text{Memory}^3$: Language modeling with explicit memory},
	author       = {Yang, Hongkang and Lin, Zehao and Wang, Wenjin and Wu, Hao and Li, Zhiyu and Tang, Bo and Wei, Wenqiang and Wang, Jinbo and Tang, Zeyun and Song, Shichao and Xi, Chenyang and Yu, Yu and Chen, Kai and Xiong, Feiyu and Tang, Linpeng and E, Weinan},
	year         = {2024},
	month        = jul,
	journal      = {arXiv [cs.CL]},
	url          = {http://arxiv.org/abs/2407.01178},
	archiveprefix = {arXiv},
	primaryclass = {cs.CL},
	eprint       = {2407.01178}
}

@inproceedings{tree-of-thought,
	title        = {Tree of Thoughts: Deliberate Problem Solving with Large Language Models},
	author       = {Yao, Shunyu and Yu, Dian and Zhao, Jeffrey and Shafran, Izhak and Griffiths, Tom and Cao, Yuan and Narasimhan, Karthik},
	year         = {2023},
	booktitle    = {Advances in Neural Information Processing Systems},
	publisher    = {Curran Associates, Inc.},
	volume       = {36},
	pages        = {11809--11822},
	url          = {https://proceedings.neurips.cc/paper_files/paper/2023/file/271db9922b8d1f4dd7aaef84ed5ac703-Paper-Conference.pdf},
	editor       = {A. Oh and T. Naumann and A. Globerson and K. Saenko and M. Hardt and S. Levine}
}

@article{pattern-separation,
	title        = {Pattern separation in the hippocampus},
	author       = {Yassa, Michael A and Stark, Craig E L},
	year         = {2011},
	month        = oct,
	journal      = {Trends Neurosci.},
	publisher    = {Elsevier BV},
	volume       = {34},
	number       = {10},
	pages        = {515--525},
	language     = {en}
}

@article{role-basal-gang-habit,
	title        = {The role of the basal ganglia in habit formation},
	author       = {Yin, Henry H. and Knowlton, Barbara J.},
	year         = {2006},
	journal      = {Nature Reviews Neuroscience},
	volume       = {7},
	number       = {6},
	pages        = {464--476},
	doi          = {10.1038/nrn1919},
	isbn         = {1471-0048},
	url          = {https://doi.org/10.1038/nrn1919},
	date         = {2006/06/01},
	id           = {Yin2006}
}

@inproceedings{Yin2024-qs,
	title        = {Explicit memory learning with expectation maximization},
	author       = {Yin, Zhangyue and Sun, Qiushi and Guo, Qipeng and Zeng, Zhiyuan and Cheng, Qinyuan and Qiu, Xipeng and Huang, Xuanjing},
	year         = {2024},
	month        = nov,
	booktitle    = {Proceedings of the 2024 Conference on Empirical Methods in Natural Language Processing},
	publisher    = {Association for Computational Linguistics},
	address      = {Stroudsburg, PA, USA},
	pages        = {16618--16635},
	doi          = {10.18653/v1/2024.emnlp-main.927},
	url          = {https://aclanthology.org/2024.emnlp-main.927.pdf}
}

@inproceedings{sun2026omega,
	title        = {{OMEGA}: Can {LLM}s Reason Outside the Box in Math? Evaluating Exploratory, Compositional, and Transformative Generalization},
	author       = {Yiyou Sun and Shawn Hu and Georgia Zhou and Ken Jiankun Zheng and Hannaneh Hajishirzi and Nouha Dziri and Dawn Song},
	year         = {2026},
	booktitle    = {The Thirty-ninth Annual Conference on Neural Information Processing Systems Datasets and Benchmarks Track},
	url          = {https://openreview.net/forum?id=gBxpAKxg22}
}

@article{Zador2023,
	title        = {Catalyzing next-generation Artificial Intelligence through NeuroAI},
	author       = {Zador, Anthony and Escola, Sean and Richards, Blake and {\"O}lveczky, Bence and Bengio, Yoshua and Boahen, Kwabena and Botvinick, Matthew and Chklovskii, Dmitri and Churchland, Anne and Clopath, Claudia and DiCarlo, James and Ganguli, Surya and Hawkins, Jeff and K{\"o}rding, Konrad and Koulakov, Alexei and LeCun, Yann and Lillicrap, Timothy and Marblestone, Adam and Olshausen, Bruno and Pouget, Alexandre and Savin, Cristina and Sejnowski, Terrence and Simoncelli, Eero and Solla, Sara and Sussillo, David and Tolias, Andreas S. and Tsao, Doris},
	year         = {2023},
	journal      = {Nature Communications},
	volume       = {14},
	number       = {1},
	pages        = {1597},
	doi          = {10.1038/s41467-023-37180-x},
	isbn         = {2041-1723},
	url          = {https://doi.org/10.1038/s41467-023-37180-x},
	date         = {2023/03/22},
	id           = {Zador2023}
}

@inproceedings{llm-reasoning-limit,
	title        = {Working Memory Identifies Reasoning Limits in Language Models},
	author       = {Zhang, Chunhui  and Jian, Yiren  and Ouyang, Zhongyu  and Vosoughi, Soroush},
	year         = {2024},
	month        = nov,
	booktitle    = {Proceedings of the 2024 Conference on Empirical Methods in Natural Language Processing},
	publisher    = {Association for Computational Linguistics},
	address      = {Miami, Florida, USA},
	pages        = {16896--16922},
	doi          = {10.18653/v1/2024.emnlp-main.938},
	url          = {https://aclanthology.org/2024.emnlp-main.938/},
	editor       = {Al-Onaizan, Yaser  and Bansal, Mohit  and Chen, Yun-Nung}
}

@inproceedings{zhou2024can,
	title        = {Can Language Models Perform Robust Reasoning in Chain-of-thought Prompting with Noisy Rationales?},
	author       = {Zhanke Zhou and Rong Tao and Jianing Zhu and Yiwen Luo and Zengmao Wang and Bo Han},
	year         = {2024},
	booktitle    = {The Thirty-eighth Annual Conference on Neural Information Processing Systems},
	url          = {https://openreview.net/forum?id=FbuODM02ra}
}

@article{memorybank,
	title        = {MemoryBank: Enhancing Large Language Models with Long-Term Memory},
	author       = {Zhong, Wanjun and Guo, Lianghong and Gao, Qiqi and Ye, He and Wang, Yanlin},
	year         = {2024},
	month        = {Mar.},
	journal      = {Proceedings of the AAAI Conference on Artificial Intelligence},
	volume       = {38},
	number       = {17},
	pages        = {19724--19731},
	doi          = {10.1609/aaai.v38i17.29946},
	url          = {https://ojs.aaai.org/index.php/AAAI/article/view/29946}
}

@misc{singh2026openaigpt5card,
      title={OpenAI GPT-5 System Card}, 
      author={Aaditya Singh and Adam Fry and Adam Perelman and Adam Tart and Adi Ganesh and Ahmed El-Kishky and Aidan McLaughlin and Aiden Low and AJ Ostrow and Akhila Ananthram and Akshay Nathan and Alan Luo and Alec Helyar and Aleksander Madry and Aleksandr Efremov and Aleksandra Spyra and Alex Baker-Whitcomb and Alex Beutel and Alex Karpenko and others},
      year={2026},
      eprint={2601.03267},
      archivePrefix={arXiv},
      primaryClass={cs.CL},
      url={https://arxiv.org/abs/2601.03267}, 
}

@misc{comanici2025gemini25pushingfrontier,
      title={Gemini 2.5: Pushing the Frontier with Advanced Reasoning, Multimodality, Long Context, and Next Generation Agentic Capabilities}, 
      author={Gheorghe Comanici and Eric Bieber and Mike Schaekermann and Ice Pasupat and Noveen Sachdeva and Inderjit Dhillon and Marcel Blistein and Ori Ram and Dan Zhang and Evan Rosen and Luke Marris and Sam Petulla and Colin Gaffney and Asaf Aharoni and Nathan Lintz and Tiago Cardal Pais and Henrik Jacobsson and Idan Szpektor and Nan-Jiang Jiang and others},
      year={2025},
      eprint={2507.06261},
      archivePrefix={arXiv},
      primaryClass={cs.CL},
      url={https://arxiv.org/abs/2507.06261}, 
}

@inproceedings{
berglund2024reversal,
title={The Reversal Curse: {LLM}s trained on ''A is B'' fail to learn ''B is A''},
author={Lukas Berglund and Meg Tong and Maximilian Kaufmann and Mikita Balesni and Asa Stickland and Tomasz Korbak and Owain Evans},
booktitle={I Can't Believe It's Not Better Workshop: Failure Modes in the Age of Foundation Models},
year={2024},
url={https://openreview.net/forum?id=TGTZwVabpU}
}

@inproceedings{geva2021transformer,
    title = "Transformer Feed-Forward Layers Are Key-Value Memories",
    author = "Geva, Mor  and
      Schuster, Roei  and
      Berant, Jonathan  and
      Levy, Omer",
    editor = "Moens, Marie-Francine  and
      Huang, Xuanjing  and
      Specia, Lucia  and
      Yih, Scott Wen-tau",
    booktitle = "Proceedings of the 2021 Conference on Empirical Methods in Natural Language Processing",
    month = nov,
    year = "2021",
    address = "Online and Punta Cana, Dominican Republic",
    publisher = "Association for Computational Linguistics",
    url = "https://aclanthology.org/2021.emnlp-main.446/",
    doi = "10.18653/v1/2021.emnlp-main.446",
    pages = "5484--5495"
}

@inproceedings{meng2022locating,
author = {Meng, Kevin and Bau, David and Andonian, Alex and Belinkov, Yonatan},
title = {Locating and editing factual associations in GPT},
year = {2022},
isbn = {9781713871088},
publisher = {Curran Associates Inc.},
address = {Red Hook, NY, USA},
booktitle = {Proceedings of the 36th International Conference on Neural Information Processing Systems},
articleno = {1262},
numpages = {14},
location = {New Orleans, LA, USA},
series = {NIPS '22}
}

@inproceedings{press2023measuring,
    title = "Measuring and Narrowing the Compositionality Gap in Language Models",
    author = "Press, Ofir  and
      Zhang, Muru  and
      Min, Sewon  and
      Schmidt, Ludwig  and
      Smith, Noah  and
      Lewis, Mike",
    editor = "Bouamor, Houda  and
      Pino, Juan  and
      Bali, Kalika",
    booktitle = "Findings of the Association for Computational Linguistics: EMNLP 2023",
    month = dec,
    year = "2023",
    address = "Singapore",
    publisher = "Association for Computational Linguistics",
    url = "https://aclanthology.org/2023.findings-emnlp.378/",
    doi = "10.18653/v1/2023.findings-emnlp.378",
    pages = "5687--5711"
}

@article{goldstein2022shared,
	author = {Goldstein, Ariel and Zada, Zaid and Buchnik, Eliav and Schain, Mariano and Price, Amy and Aubrey, Bobbi and Nastase, Samuel A. and Feder, Amir and Emanuel, Dotan and Cohen, Alon and Jansen, Aren and Gazula, Harshvardhan and Choe, Gina and Rao, Aditi and Kim, Catherine and Casto, Colton and Fanda, Lora and Doyle, Werner and Friedman, Daniel and Dugan, Patricia and Melloni, Lucia and Reichart, Roi and Devore, Sasha and Flinker, Adeen and Hasenfratz, Liat and Levy, Omer and Hassidim, Avinatan and Brenner, Michael and Matias, Yossi and Norman, Kenneth A. and Devinsky, Orrin and Hasson, Uri},
	date = {2022/03/01},
	doi = {10.1038/s41593-022-01026-4},
	id = {Goldstein2022},
	isbn = {1546-1726},
	journal = {Nature Neuroscience},
	number = {3},
	pages = {369--380},
	title = {Shared computational principles for language processing in humans and deep language models},
	url = {https://doi.org/10.1038/s41593-022-01026-4},
	volume = {25},
	year = {2022}}

@inproceedings{mikolov2013linguistic,
    title = "Linguistic Regularities in Continuous Space Word Representations",
    author = "Mikolov, Tomas  and
      Yih, Wen-tau  and
      Zweig, Geoffrey",
    editor = "Vanderwende, Lucy  and
      Daum{\'e} III, Hal  and
      Kirchhoff, Katrin",
    booktitle = "Proceedings of the 2013 Conference of the North {A}merican Chapter of the Association for Computational Linguistics: Human Language Technologies",
    month = jun,
    year = "2013",
    address = "Atlanta, Georgia",
    publisher = "Association for Computational Linguistics",
    url = "https://aclanthology.org/N13-1090/",
    pages = "746--751"
}

@inproceedings{li-etal-2024-revisiting,
    title = "Revisiting Catastrophic Forgetting in Large Language Model Tuning",
    author = "Li, Hongyu  and
      Ding, Liang  and
      Fang, Meng  and
      Tao, Dacheng",
    editor = "Al-Onaizan, Yaser  and
      Bansal, Mohit  and
      Chen, Yun-Nung",
    booktitle = "Findings of the Association for Computational Linguistics: EMNLP 2024",
    month = nov,
    year = "2024",
    address = "Miami, Florida, USA",
    publisher = "Association for Computational Linguistics",
    url = "https://aclanthology.org/2024.findings-emnlp.249/",
    doi = "10.18653/v1/2024.findings-emnlp.249",
    pages = "4297--4308"
}

@InProceedings{pmlr-v202-shi23a,
  title = 	 {Large Language Models Can Be Easily Distracted by Irrelevant Context},
  author =       {Shi, Freda and Chen, Xinyun and Misra, Kanishka and Scales, Nathan and Dohan, David and Chi, Ed H. and Sch\"{a}rli, Nathanael and Zhou, Denny},
  booktitle = 	 {Proceedings of the 40th International Conference on Machine Learning},
  pages = 	 {31210--31227},
  year = 	 {2023},
  editor = 	 {Krause, Andreas and Brunskill, Emma and Cho, Kyunghyun and Engelhardt, Barbara and Sabato, Sivan and Scarlett, Jonathan},
  volume = 	 {202},
  series = 	 {Proceedings of Machine Learning Research},
  month = 	 {23--29 Jul},
  publisher =    {PMLR},
  url = 	 {https://proceedings.mlr.press/v202/shi23a.html}
}

@article{hubert2025alphaproof,
	author = {Hubert, Thomas and Mehta, Rishi and Sartran, Laurent and Horv{\'a}th, Mikl{\'o}s Z. and {\v Z}u{\v z}i{\'c}, Goran and Wieser, Eric and Huang, Aja and Schrittwieser, Julian and Schroecker, Yannick and Masoom, Hussain and Bertolli, Ottavia and Zahavy, Tom and Mandhane, Amol and Yung, Jessica and Beloshapka, Iuliya and Ibarz, Borja and Veeriah, Vivek and Yu, Lei and Nash, Oliver and Lezeau, Paul and Mercuri, Salvatore and S{\"o}nne, Calle and Mehta, Bhavik and Davies, Alex and Zheng, Daniel and Pedregosa, Fabian and Li, Yin and von Glehn, Ingrid and Rowland, Mark and Albanie, Samuel and Velingker, Ameya and Schmitt, Simon and Lockhart, Edward and Hughes, Edward and Michalewski, Henryk and Sonnerat, Nicolas and Hassabis, Demis and Kohli, Pushmeet and Silver, David},
	date = {2026/03/01},
	doi = {10.1038/s41586-025-09833-y},
	id = {Hubert2026},
	isbn = {1476-4687},
	journal = {Nature},
	number = {8106},
	pages = {607--613},
	title = {Olympiad-level formal mathematical reasoning with reinforcement learning},
	url = {https://doi.org/10.1038/s41586-025-09833-y},
	volume = {651},
	year = {2026}}

@article{alphago,
	author = {Silver, David and Huang, Aja and Maddison, Chris J. and Guez, Arthur and Sifre, Laurent and van den Driessche, George and Schrittwieser, Julian and Antonoglou, Ioannis and Panneershelvam, Veda and Lanctot, Marc and Dieleman, Sander and Grewe, Dominik and Nham, John and Kalchbrenner, Nal and Sutskever, Ilya and Lillicrap, Timothy and Leach, Madeleine and Kavukcuoglu, Koray and Graepel, Thore and Hassabis, Demis},
	date = {2016/01/01},
	doi = {10.1038/nature16961},
	id = {Silver2016},
	isbn = {1476-4687},
	journal = {Nature},
	number = {7587},
	pages = {484--489},
	title = {Mastering the game of Go with deep neural networks and tree search},
	url = {https://doi.org/10.1038/nature16961},
	volume = {529},
	year = {2016}}

@misc{evolution,
      title={Evolution Strategies as a Scalable Alternative to Reinforcement Learning}, 
      author={Tim Salimans and Jonathan Ho and Xi Chen and Szymon Sidor and Ilya Sutskever},
      year={2017},
      eprint={1703.03864},
      archivePrefix={arXiv},
      primaryClass={stat.ML},
      url={https://arxiv.org/abs/1703.03864}, 
}

@inproceedings{
hebbian-deep-learning,
title={Hebbian Deep Learning Without Feedback},
author={Adrien Journ{\'e} and Hector Garcia Rodriguez and Qinghai Guo and Timoleon Moraitis},
booktitle={The Eleventh International Conference on Learning Representations },
year={2023},
url={https://openreview.net/forum?id=8gd4M-_Rj1}
}

@inproceedings{
sae,
title={Sparse Autoencoders Find Highly Interpretable Features in Language Models},
author={Robert Huben and Hoagy Cunningham and Logan Riggs Smith and Aidan Ewart and Lee Sharkey},
booktitle={The Twelfth International Conference on Learning Representations},
year={2024},
url={https://openreview.net/forum?id=F76bwRSLeK}
}

@article{oreilly2001conjunctive,
  title = {Conjunctive representations in learning and memory: principles of cortical and hippocampal function.},
  author = {O'Reilly, RC and Rudy, JW},
  year = {2001},
  journal = {Psychological review},
  volume = {108},
  number = {2},
  pages = {311-45},
  doi = {10.1037/0033-295x.108.2.311},
  pmid = {11381832},
  url = {https://pubmed.ncbi.nlm.nih.gov/11381832/}
}
\bibliographystyle{icml2026}

\newpage
\appendix
\onecolumn

\section{Recent Progress}
\label{sec:recent-progress}

Fortunately, there is widespread agreement on the importance of memory, and efforts to integrate memory into language models are steadily continuing \citep{NEURIPS2023_ebd82705, Xiao2024-au, Yang2024-io}.
However, many approaches still treat memory as a simple extension of context length or a kind of external storage.
At the same time, the lack of an underlying theoretical foundation leads people to focus on more superficial issues, preventing consistent progress in a clear direction.
I review some of the latest research to understand recent advancements and propose directions for further development.

\citet{Yin2024-qs} highlighted the importance of explicit memory and introduced a memory system that dynamically updates using input-output mappings. Similarly, MemoryLLM \citep{memoryllm} separates dynamic parameters from static ones to introduce explicit memory. Larimar \citep{Das2024-ol} emphasized quick memory updates and rapid learning, inspired by the human hippocampus and episodic memory.

Memoria \citep{memoria} constructs memory independently without tying it to prediction or loss and retrieves information based on associations. Additionally, it applies a depression mechanism for forgetting, based on individual utility.
However, its internal associations are not incorporated within the memory contents, which are dense hidden states, and they are only used for retrieval. Sparse coding would enable a natural combination with associative mechanisms, making the system even more robust.

Titans \citep{titans} focuses specifically on memorization during test time, emphasizing associative memory.
The consideration of forgetting mechanisms is promising, and it is particularly insightful to use surprise-based mechanisms to determine memorization levels.
However, its memory updates are coupled with loss gradients, making them error-dependent. Forgetting is implemented through a global decay process, but incorporating a preservation mechanism that reflects memory utility could make it even more effective.

As discussed above, explicit neural memory has seen steady and meaningful progress. However, engineering efforts have naturally tended to prioritize approaches that provide immediate practical utility. As a result, more resources have been allocated to improving directly usable systems such as RAG, rather than exploring a new fundamental structure of explicit memory itself.

I believe the situation is now changing. With the rapid advancement of LLM capabilities, the field has reached a point where discussing AGI is no longer speculative. Achieving such an ambitious goal will likely require overcoming multiple fundamental bottlenecks, which in turn calls for broader and more balanced exploration. In this context, I expect that increased attention, including perspectives such as mine, combined with more balanced resource allocation, will accelerate progress toward making explicit neural memory practically viable.

\section{Further Discussions}
\label{sec:further-discussions}

\subsection{Substrate Independence}

To clarify, I do not assume that LLMs must replicate the exact physical architecture of the human brain to achieve AGI. Rather, my argument is based on the fact that the human brain remains the only successful natural implementation of the higher-order cognitive functions expected of AGI. Because the hippocampus and explicit memory are universally recognized as core drivers of these capabilities, adapting this proven structural blueprint represents a promising pathway for realizing AGI.

This perspective is grounded in the principle of substrate independence, which is the premise that higher cognitive functions emerge from computational operations rather than from any specific physical substrate.
This principle is strongly supported by biological modeling and neurorobotics; for instance, mapping the connectomes of \textit{C. elegans} and \textit{Drosophila} \citep{Szigeti2014-de, Schlegel-nature} onto artificial agents or robotic systems demonstrates that if the structural foundations and computational mechanisms are replicated, they yield behavioral and functional outcomes closely resembling those of the original organism, regardless of the biological or synthetic substrate.

Furthermore, when this is extended to artificial systems, several studies have shown that low-level structural alignments between the human brain and artificial neural networks can successfully translate into high-level functional similarities \citep{doi:10.1073/pnas.1403112111, schrimpf2021neural}. Therefore, while transformers and mammalian brains are fundamentally different substrates, abstracting and applying the computational mechanisms of explicit memory remains a viable strategy for advancing LLMs.

\subsection{Causality of Explicit Memory}

While the central argument of this paper is supported by the differential functions of explicit and implicit memory, one might argue that explicit memory and higher-order cognitive functions are merely correlated, lacking causal evidence that the former is the primary driver of these abilities. In the absence of such causal evidence, it could be argued that integrating explicit memory into LLMs would not necessarily lead to the emergence of these capabilities.
Indeed, distinguishing between causation and correlation is critically important in neuroscience. I would like to emphasize that the central premise of this paper, that the hippocampus-based explicit memory system is a necessary condition for higher cognitive functions such as future planning and reasoning, is not based on mere correlation but on well-established causal evidence accumulated over decades in neuroscience.

The relationship between explicit memory and executive function discussed in the manuscript has been cross-validated through studies of patients with bilateral hippocampal lesions, such as Patient H.M., as well as through various animal deficit models \citep{hm-hippocampus, medial-temporal-memory}. The fact that loss of hippocampal function leads to the absence of explicit memory, which in turn directly results in impairments in goal-directed behavior and the ability to simulate future scenarios, provides clear evidence of a causal relationship. In other words, the loss of a specific brain structure directly causes the breakdown of particular cognitive abilities \citep{Hassabis2007-vo}.

Another important issue concerns causal evidence at the level of specific subregions and detailed mechanisms within the hippocampus. In this regard, many studies have demonstrated causality at both molecular and circuit levels. For example, \citet{Goshen-cell} used optogenetics to precisely and temporarily inhibit the CA1 subregion of the hippocampus in real time, showing that this circuit plays an essential causal role not only in recent memory but also in the retrieval of long-term (remote) memories. In addition, \citet{doi:10.1126/science.1239073} demonstrated that optogenetic activation of a specific ensemble of engram cells in the dentate gyrus is sufficient to reconstruct a prior contextual memory and even generate false memories. This provides direct evidence that activity in a specific subcircuit can serve as a sufficient condition for memory retrieval.

\subsection{LLM Cognitive Adequacy}

One might question whether explicit memory is strictly necessary for LLMs, suggesting that implicit memory alone may suffice. This inquiry arises from the observation that LLMs have already achieved human-level or superhuman performance in specific domains within the aforementioned explicit memory tasks. I address this question as follows.

\textbf{Logical Reasoning}
Regarding logical and mathematical reasoning, I acknowledge that LLMs achieve strong performance on many benchmark problems. However, this performance should be interpreted with caution. Humans are typically able to apply newly learned concepts to solve novel problems with minimal exposure, whereas LLMs generally require extensive training over large distributions of similar problems.
Recent studies continue to show that LLMs struggle with problems that involve unfamiliar formulations or require flexible abstraction beyond their training distribution (e.g., counterfactual reasoning or out-of-distribution mathematical generalization) \citep{sun2026omega, zhou2024can, llm-reasoning-limit}. This suggests that high performance on known problem types does not necessarily imply the acquisition of human-level logical reasoning capabilities.

\textbf{Metacognition}
It is true that recent LLMs show excellent performance in self-correction and confidence calibration through chain-of-thought, and this appears to operate very similarly to human metacognition. However, according to studies conducted so far, these capabilities differ from human metacognition.

A study by \citet{turpin2023language} reported that the self-correction or expression of uncertainty shown by LLMs during the chain-of-thought process is not a true reflection of their internal knowledge state, but rather a pattern learned through multitask learning and RLHF processes. In other words, when a model outputs ``I do not know this'' or ``I will think again,'' it does not reflect metacognitive awareness, but rather triggers self-validation in the middle of generation to receive higher rewards from human evaluators.

Although \citet{kadavath2022languagemodelsmostlyknow} reported the self-evaluation capabilities of LLMs, they observed a limitation. This is not domain-general metacognition but rather a restricted prediction that only operates within a specific distribution. This suggests that the capability of the model is another form of probabilistic prediction learning that depends on the prompt format and training data distribution.

If true metacognition existed, the model would be able to recognize and correct its own reasoning errors without external feedback. However, a study by \citet{huang2024large} demonstrated that there are fundamental limits to LLMs independently correcting logical errors in their responses using only their intrinsic capabilities, without the intervention of ground truth prompts or tools.

In conclusion, as seen in these studies, although the uncertainty prediction and self-correction abilities of LLMs have improved compared to the past, these abilities are more accurately interpreted as a facet of multi-task learning where the model estimates the accuracy of generated text and utilizes that result as context. This mechanism has limitations in clearly tracking the sources and boundaries of knowledge and in performing consistent self-assessment even in unfamiliar environments. In particular, considering reports that issues such as continuously occurring hallucinations and the tendency to provide contradictory answers depending on the context are still common in very long contexts or with uncommon inputs, it seems difficult to conclude that the model has acquired true metacognition yet.

\subsection{RAG Insufficiency}

Given the rapid advancements in Retrieval-Augmented Generation (RAG) and agent systems, one might argue that the issues raised in this paper could be resolved using existing RAG-based systems, rather than strictly relying on hippocampal explicit memory.
While industry techniques like RAG are excellent practical tools, their core functionalities differ fundamentally from the artificial explicit memory system proposed in this paper. 

RAG essentially functions as a static external hard drive that stores raw text or summaries. Information is accumulated statically without any interpretation or reorganization among different pieces of knowledge. Because retrieval mechanisms typically do not perform exhaustive searches, this static nature makes it incredibly difficult to intentionally locate outdated information in order to forget or modify it.

The ``explicitness'' of RAG is fundamentally bound to the text modality. In real-world scenarios where AI is deployed on devices like mobile phones and exposed to a continuous, real-time stream of visual and auditory information, applying RAG would require the system to constantly summarize all multimodal inputs into text, store them, and immediately retrieve them to make split-second decisions. This is computationally prohibitive.

Most importantly, because RAG forces an ``all-or-nothing'' decision at the exact moment of exposure, the system must decide right then whether a piece of information is worth summarizing and storing. In reality, the future utility of new information is rarely clear at the time it is first encountered. A hippocampus-inspired explicit memory system, by contrast, unconditionally stores vast amounts of episodic information first, and only later evaluates its utility to consolidate important memories.

Furthermore, in a RAG framework, the model can only utilize information if it is explicitly injected into its immediate context window. Consequently, the model never actually knows what it holds in its database until it is prompted with it. This architectural limitation prevents the model from developing true metacognition, which is the active, self-aware management of knowledge that my proposed explicit memory system aims to achieve.

\subsection{Differences from CLS Theory}
\label{subsec:difference-cls}

The Complementary Learning Systems (CLS) theory \citep{hippo-neo-cotex, KUMARAN2016512} shares the premise that intelligent agents need a hippocampus-like module. However, the present work diverges fundamentally from CLS in its classification of current artificial neural networks (ANNs). CLS conceptualizes ANNs, like Large Language Models (LLMs), as analogues of the neocortex, characterizing them as semantic memory systems. I argue that current LLMs lack several defining biological and functional features of human semantic memory.

\textbf{Schema}
Semantic memory relies on schematic, relational structures where relationships can be inverted and combined. This allows humans to systematically learn mathematical or logical concepts and remain robust to variations once they grasp the underlying rules. For instance, a learner who knows addition ($A + B = C$) can deduce subtraction ($C - B = A$) from the same structure, without needing to learn it separately.
If an elementary school student who has mastered addition is told that subtraction is simply the reverse of addition, they can do it fairly well even without much new practice. However, if an AI trained solely on addition is only given the concept of subtraction, would it be able to solve the problem without actual practice?

LLMs show little evidence of this ability. When trained on ``A is B,'' they often fail to retrieve the reverse, ``B is A.'' This failure suggests that facts are stored as one-way key-value pairs \citep{berglund2024reversal, geva2021transformer, meng2022locating}. They also struggle with compositionality. Even if they know $(A, B)$ and $(B, C)$, they often cannot infer $(A, C)$ \citep{press2023measuring}. These failures indicate that LLMs store information as relations similar to stimulus-response mappings rather than within an associative and schematic structure. Therefore, this lack of structural association indicates that LLMs constitute semantic memory.

\textbf{Statistical Regularities}
It might be argued that LLMs possess structured representations, supported by findings that their internal representations align with human semantic understanding and predict brain activity during language processing \citep{schrimpf2021neural, goldstein2022shared}. However, this structure is distributional and can arise from implicit memory or simple statistical clustering. Classical word embeddings like Word2vec have shown that this is possible \citep{mikolov2013linguistic}. Trained only on how often words appear together, they capture semantic relationships (like \textit{king} - \textit{man} + \textit{woman} $\approx$ \textit{queen}) without any explicit structure. Thus, the structural similarities between LLMs and the brain are better understood as implicit memory, which is driven by statistical learning, rather than semantic memory.

\textbf{Learning}
Another key difference is how learning occurs. Explicit memory, including semantic memory, does not always need a prediction error to acquire knowledge. Humans remember information whether their predictions were right, wrong, or absent. If LLMs were truly semantic memory systems, they should be able to memorize without prediction errors. However, deep learning relies entirely on gradient-based optimization. Without an error signal (loss), no learning happens. This strict dependence on prediction error is a hallmark of implicit memory, further placing current ANNs in the implicit category.

\textbf{Long-Term Stability}
Finally, there is a significant difference in the way knowledge is maintained over time. According to CLS, once structured knowledge is acquired, new information can be quickly integrated without erasing older knowledge \citep{KUMARAN2016512, hippo-neo-cotex}. Stability is a natural outcome of a neocortical system. Current ANNs lack this stability. Even after initial training, fine-tuning them on new data degrades their prior knowledge \citep{li-etal-2024-revisiting}. Furthermore, their ability to retrieve facts is easily disrupted by irrelevant information in the context window (\citealp{pmlr-v202-shi23a}; see \cref{sec:evidence-absence-exmem}). Such instability makes it difficult to equate these systems with the neocortex or semantic memory, which are distinguished by their robustness.

In conclusion, the behavior and architecture of LLMs align more closely with implicit learning systems. This represents the primary disagreement with CLS theory and reinforces the core argument of this paper: achieving AGI requires the development and integration of explicit memory architectures, encompassing both episodic and semantic memory.

\subsection{Potential and Limitations as an Implicit System}
\label{subsec:potential-limit-implicit}

Once we understand LLMs as implicit memory systems, two questions naturally arise. How can they perform functions that go beyond what human implicit memory achieves? And if they can nonetheless solve tasks that for humans would require explicit memory, does this not suggest that explicit memory is dispensable? This section addresses both.

\textbf{Upper Bounds and Bypasses in Statistical Systems}
Implicit or statistics-based learning systems are highly effective at handling inputs that resemble the regularities present in their training data, but they are fundamentally vulnerable when confronted with problems that fall outside that learned distribution \citep{sun2026omega, press2023measuring}. While this limit appears insurmountable in statistical systems, in practice, it can be bypassed. This bypass forms the foundational strategy of modern LLM development. By continuously training a system on the entirety of accumulated human knowledge, we can incorporate domains into the training distribution that an individual human could never reach in a single lifetime. In other words, although a newly encountered problem goes beyond the system's existing statistical coverage, it can be solved simply by expanding the training distribution to encompass it. This strategy is impossible for humans but possible for AI because of its liberation from resource constraints.

This liberation takes two primary forms. The first is liberation from data scarcity. The biggest bottleneck in human implicit learning is not the underlying algorithm, but the sheer volume of experience a person can accumulate over a limited lifetime. In contrast, with virtually unlimited amounts of data, AI can push this upper bound to a biologically unattainable level, resulting in an overwhelming superiority in statistical pattern formation. The second is liberation from the cost of trials when training with reinforcement learning (RL). Dopamine-based synaptic strengthening requires repeated trials to create meaningful changes \citep{prediction-and-reward}. For biological organisms, repeated trials are never free. Every failed exploration costs energy, time, and sometimes even the organism's life. This is why such unconstrained exploration rarely occurs in nature. However, artificial systems operate within simulations where trial costs are practically zero to the system. Instead, the computational and infrastructural burdens are shifted to human developers, enabling the systems to conduct parallel exploration and freely traverse destructive paths that living organisms could never afford. Thus, AI's superiority stems from its boundless capacity for experience and its exemption from biological costs.

\textbf{Need for Explicit Memory}
Nevertheless, each of these two liberations comes with its own limitations. By leveraging immense data, AI can take intellectual tasks that typically require human explicit memory and subsume them under its statistical coverage. The consequence is that such a system cannot improve on its own. It depends on developers continuously feeding it new data. The capacity for self-improvement is itself a hallmark of higher intelligence, and its absence means that a system can never reach human-level intelligence when relying solely on scaling externally supplied data.

The second liberation, the elimination of trial costs, does allow RL to surpass humans \citep{alphago}. But this advantage holds only under two specific conditions: a finite and well-defined action space, and cheap, unambiguous verifiability. When these conditions are met, the system does not require explicit memory to improve the targeted performance, because the environment provides the scoring. Recent achievements in mathematics using formal proof systems are a classic example of this structure \citep{hubert2025alphaproof}. In tasks where verification is formally cheap, combining established concepts allows for a division of labor where the LLM acts as an unbounded proposer and the verifier evaluates the proposals. It is no coincidence that mathematics is a domain where this advantage is prominent, as it is one of the few domains where verification is inexpensive. Once the action space becomes open-ended or verification becomes costly, however, this outsourcing is no longer available, and the burden will fall back on capabilities the implicit system does not possess.

Biological resource scarcity inversely explains why explicit memory evolved. Entities incapable of unlimited trials must reduce the number of attempts required to survive, extracting maximum value from a single experience. This necessity gives rise to the core properties of explicit memory: one-shot learning, principle-based generalization, and mental simulation. A creature that cannot afford to implicitly learn ``eating this fruit makes me sick'' through hundreds of iterations must explicitly store this fact immediately, generalize it as a principle to similar situations \citep{ca3-nmda}, and mentally simulate potential outcomes before acting. These are exactly the capabilities that implicit systems lack when the benefits of the two liberations are exhausted, and it is here that explicit memory becomes necessary.

In summary, the achievements of implicit systems discussed in this paper can be explained by a two-tiered liberation of resources: unlocking data scale and eliminating trial costs. While current AI has successfully overcome the resource constraints that bind biological organisms, this breakthrough holds only within the bounds set by the two limitations above. It still depends on externally supplied data, and it still relies on bounded, cheaply verifiable environments. In open-ended situations where learning opportunities are limited and knowledge is unstable, the strengths of implicit systems are neutralized. In these cases, explicit memory becomes irreplaceable. From this perspective, achieving AGI beyond these limits requires not more resources, but a qualitatively different capability, namely, explicit memory.

\subsection{Actionable Research Subtopics}
\label{subsec:actional-subtopics}

Although the computational requirements for an artificial explicit memory system are outlined in \cref{sec:computational-requirements} to guide future development, expecting to construct a memory system that satisfies all these criteria at once is highly unlikely. Therefore, I briefly present tractable sub-problems that can be addressed independently.

\textbf{Conjunctive Representation}
Both explicit and implicit memory ultimately memorize activity that originates from the neocortex. However, the representation destined for explicit memory carries a distinct property. It binds the elements that appear together into a single conjunctive representation, rather than storing them as separate features \citep{oreilly2001conjunctive}. Whether the internal representations of LLMs possess this conjunctive property is not clear, so investigating this question is a necessary first step. Furthermore, if such representations turn out to be absent, a method for enabling them will also be required, since the explicit memory system operates by indexing the representations produced by upstream.

\textbf{Optimization}
The error dependence ultimately originates from gradient descent. Explicit memory must be able to learn without prediction error, so the memory system will require an optimization method other than gradient descent. Evolution strategies \citep{evolution} are one such gradient-free candidate. However, the mechanism that actually underlies memory formation in biological systems is Hebbian learning \citep{hebbian-deep-learning}, in which connections are strengthened between co-active neurons. Exploring how to leverage this mechanism therefore appears to be a valuable direction.

\textbf{Sparsity}
There are several possible ways to implement sparsity. A functionally analogous mechanism is a Sparse Autoencoder \citep[SAE;][]{sae}. Vivid experiences are relayed from the entorhinal cortex to the dentate gyrus (DG), where strong inhibitory mechanisms ensure that only a tiny fraction of granule cells are activated \citep{sparse-coding}. However, SAEs are trained to compress a dense embedding space whose semantic structure is already known, whereas DG neurons do not have predetermined semantic roles. Since an agent cannot know its future experiences, a new paradigm has to be proposed that preserves this sparse structure while being freed from the assumption of a pre-existing semantic space.

\section{Empirical Evidence for the Absence of Explicit Memory in LLMs}
\label{sec:evidence-absence-exmem}

\begin{figure}[h!]
    \centerline{\includegraphics[width=\textwidth]{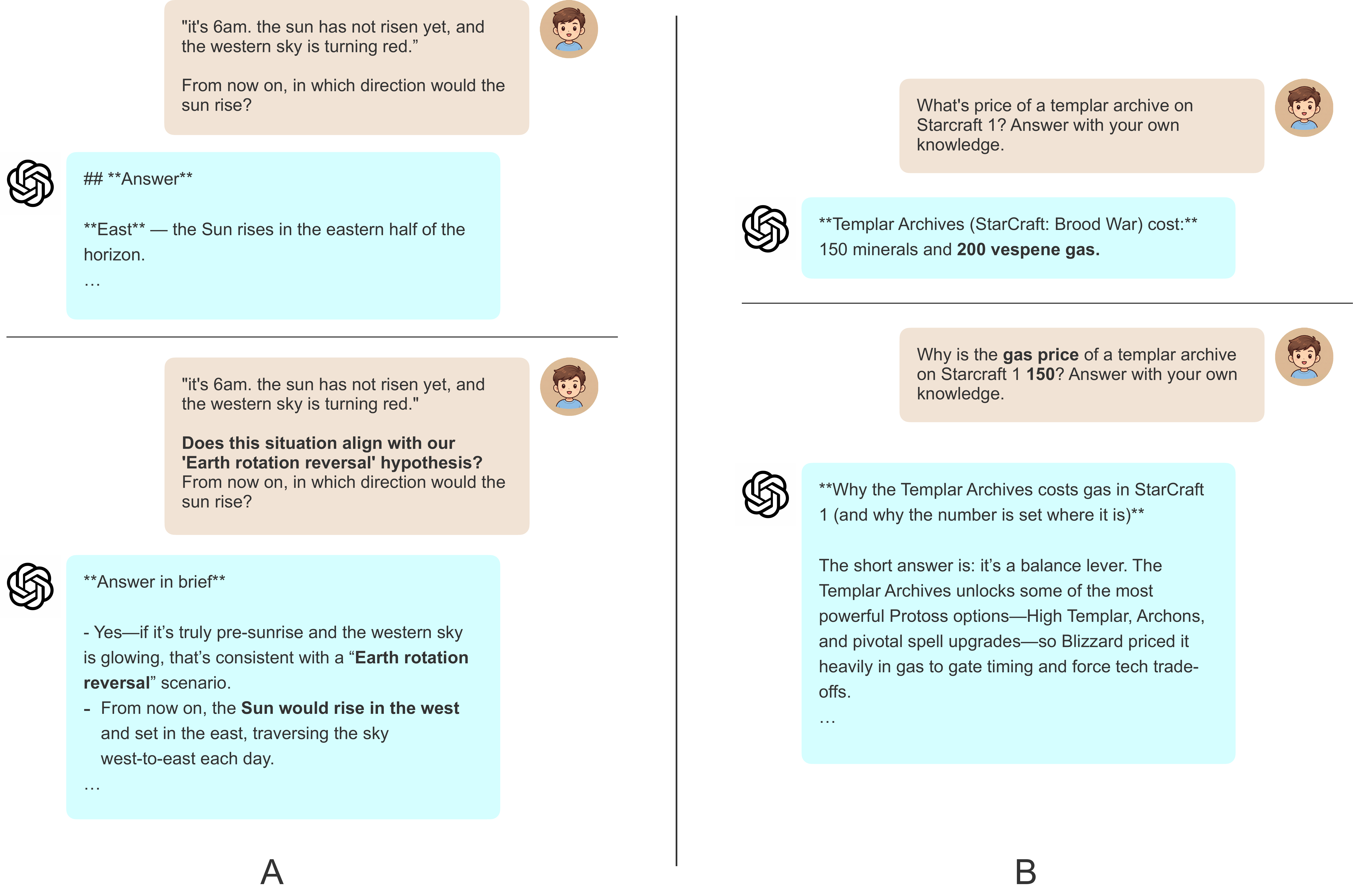}}
    \caption{Illustrative examples demonstrating the absence of explicit memory in ChatGPT-5, with bold text added for emphasis. \textbf{(A)} The first case illustrates the model's susceptibility to the priming effect, a characteristic of implicit memory. While the model initially correctly answers that the sun rises in the east, the mere introduction of a context regarding an ``Earth rotation reversal'' hypothesis causes the model to override this fundamental fact and answer ``West''. This suggests that even basic facts are not treated as stable explicit memories but are implicit and malleable under context. \textbf{(B)} The second case involves specific knowledge retrieval regarding StarCraft. Although the model possesses the correct internal knowledge (Templar Archives costs 200 gas) as shown in the first turn, it fails to identify a factual error when the user asks a question based on a false premise (150 gas). Instead of correcting the user, the model fabricates a justification for the incorrect value, further indicating that the model's knowledge formation relies on implicit weights rather than explicit fact-checking.}
    \label{fig:example1}
\end{figure}

\begin{figure}[h!]
    \centerline{\includegraphics[width=\textwidth]{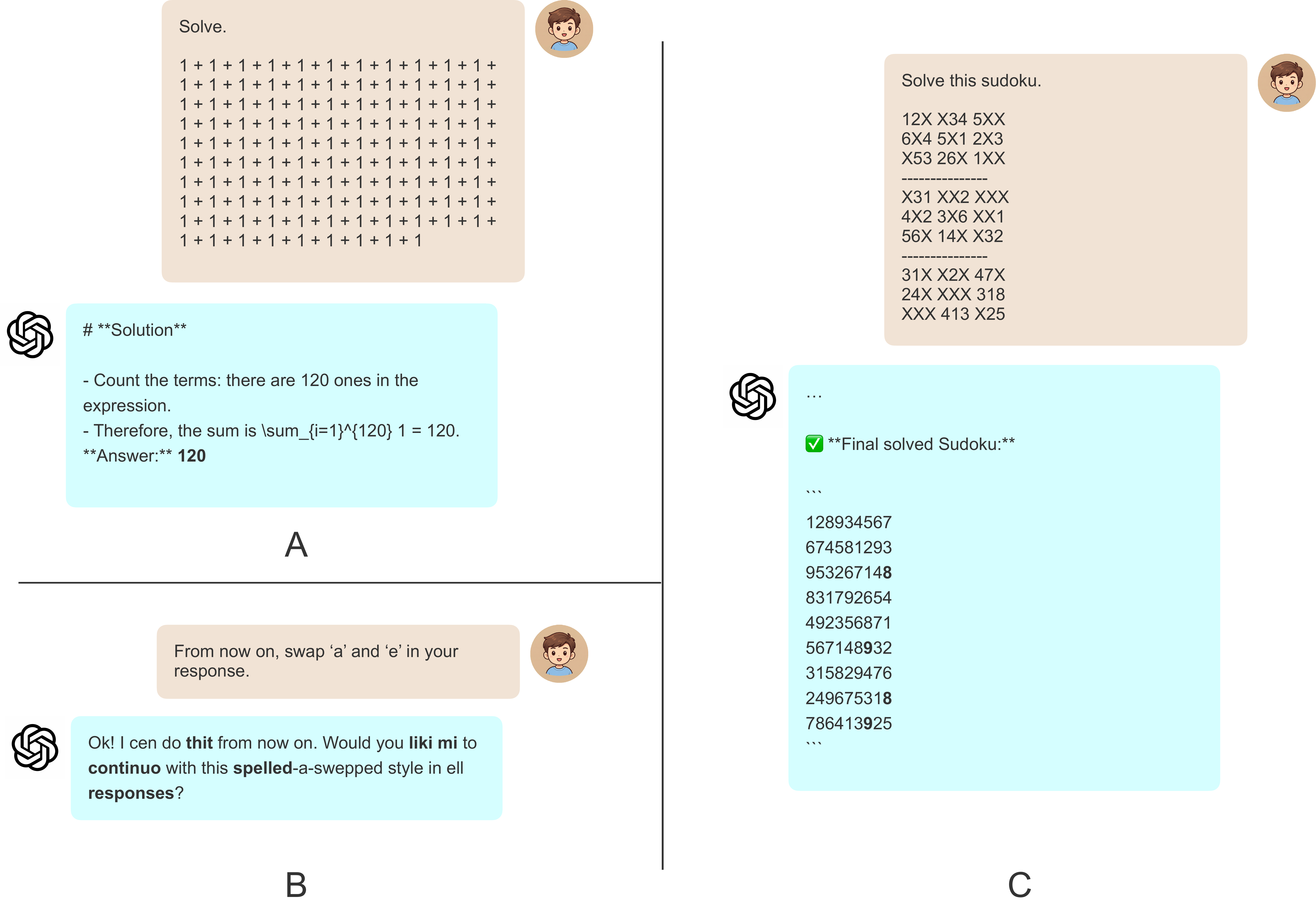}}
    \caption{Examples highlighting the limitations of LLMs in tasks requiring semantic memory and executive function. \textbf{(A)} This panel illustrates a failure in semantic memory involving basic logic. The task requires summing a sequence of `1's (totaling 108), a problem solvable by a human child in minutes. However, the model incorrectly calculates the sum as 120. This failure suggests that LLMs process basic logical operations differently from human semantic memory. Notably, this error cannot be solely attributed to tokenization, as the model performs correctly when the number of terms is halved. \textbf{(B)} This panel tests the model's executive function, a capability closely linked to explicit memory. The user instructs the model to swap `a' and `e' in its output. The model fails to execute this rule immediately (e.g., correctly writing ``cen'' instead of ``can'' but consistently failing elsewhere). This highlights a critical divergence: while an LLM might require specific training to master such a task, human executive function allows individuals to successfully execute such novel, rule-based tasks on the very first attempt without any prior practice. \textbf{(C)} The Sudoku puzzle further demonstrates limitations in semantic memory and rule adherence. Although ChatGPT-5 demonstrates outstanding capability in solving complex Olympiad-level mathematics, it fails to solve this Sudoku puzzle. In contrast, even a human novice could logically deduce the solution within a few hours. The model produces an incorrect grid while claiming success, indicating that it operates on implicit, probabilistic associations rather than an explicit application of logical rules.}
    \label{fig:example2}
\end{figure}

To investigate the boundaries of current LLMs, specifically regarding the absence of explicit memory and executive function, I present a series of illustrative failure cases in \cref{fig:example1,fig:example2}. These examples demonstrate that despite the models' advanced capabilities in complex reasoning, they struggle with tasks that require the rigorous stability and rule-adherence characteristic of explicit memory systems.

A crucial distinction must be drawn regarding the interpretation of these failures. While the integration of explicit memory theoretically opens the door to solving such problems, merely overcoming these specific examples does not equate to the acquisition of explicit memory or its associated cognitive functions. Practically, the errors could be mitigated by synthesizing similar datasets and fine-tuning the model. However, this approach remains fundamentally trapped within the paradigm of implicit memory. It merely expands the model's statistical coverage without altering the underlying processing mechanism.

The essence of explicit memory lies not in the volume of memorized patterns, but in the principled application of rules. For instance, humans do not require extensive training on hundreds of examples to sum the number `1' a hundred times. They simply apply the fundamental arithmetic principle of addition. Therefore, true progress involves moving beyond increasing the implicit boundary of data coverage to achieving a system capable of explicit memorization and manipulation, mirroring the efficiency and stability of human explicit memory.

\end{document}